\documentclass[runningheads]{llncs}

\usepackage{eccv}

\usepackage{eccvabbrv}

\usepackage{graphicx}
\usepackage{booktabs}
\usepackage{bm}

\usepackage[accsupp]{axessibility}  % Improves PDF readability for those with disabilities.

\usepackage{hyperref}

\usepackage{orcidlink}

\begin{document}

% ---------------------------------------------------------------
% TODO REVIEW: Replace with your title
\title{In-context Region-based Drag: Drag Any Region to Any Shape} 

% TODO REVIEW: If the paper title is too long for the running head, you can set
% an abbreviated paper title here. If not, comment out.
% \titlerunning{Abbreviated paper title}

% TODO FINAL: Replace with your author list. 
% Include the authors' OCRID for the camera-ready version, if at all possible.

% Equal contribution mark
\newcommand{\equalcontrib}{\textsuperscript{\textdagger}}
\newcommand{\corresponding}{\textsuperscript{*}}

\author{Jiacheng Sui\equalcontrib \and
Tianyu Hao\orcidlink{0009-0000-6488-5985}\equalcontrib \and
Bingjie Gao\orcidlink{0000-0003-3622-7509} \and
Li Niu\orcidlink{0000-0003-1970-8634}\corresponding \and
Guangtao Zhai\orcidlink{0000-0001-8165-9322}}

% TODO FINAL: Replace with an abbreviated list of authors.
\authorrunning{J. Sui et al.}
% First names are abbreviated in the running head.
% If there are more than two authors, 'et al.' is used.

% TODO FINAL: Replace with your institution list.
\institute{
Shanghai Jiao Tong University\\
\email{ jcsui01@sjtu.edu.cn, htianyu429@gmail.com, whynothaha@sjtu.edu.cn, ustcnewly@sjtu.edu.cn, zhaiguangtao@sjtu.edu.cn }
}

\maketitle

\begingroup
\renewcommand\thefootnote{\textdagger}
\footnotetext{Equal contribution.}
\renewcommand\thefootnote{*}
\footnotetext{Corresponding author.}
\endgroup

\begin{center}
    \includegraphics[width=\textwidth]{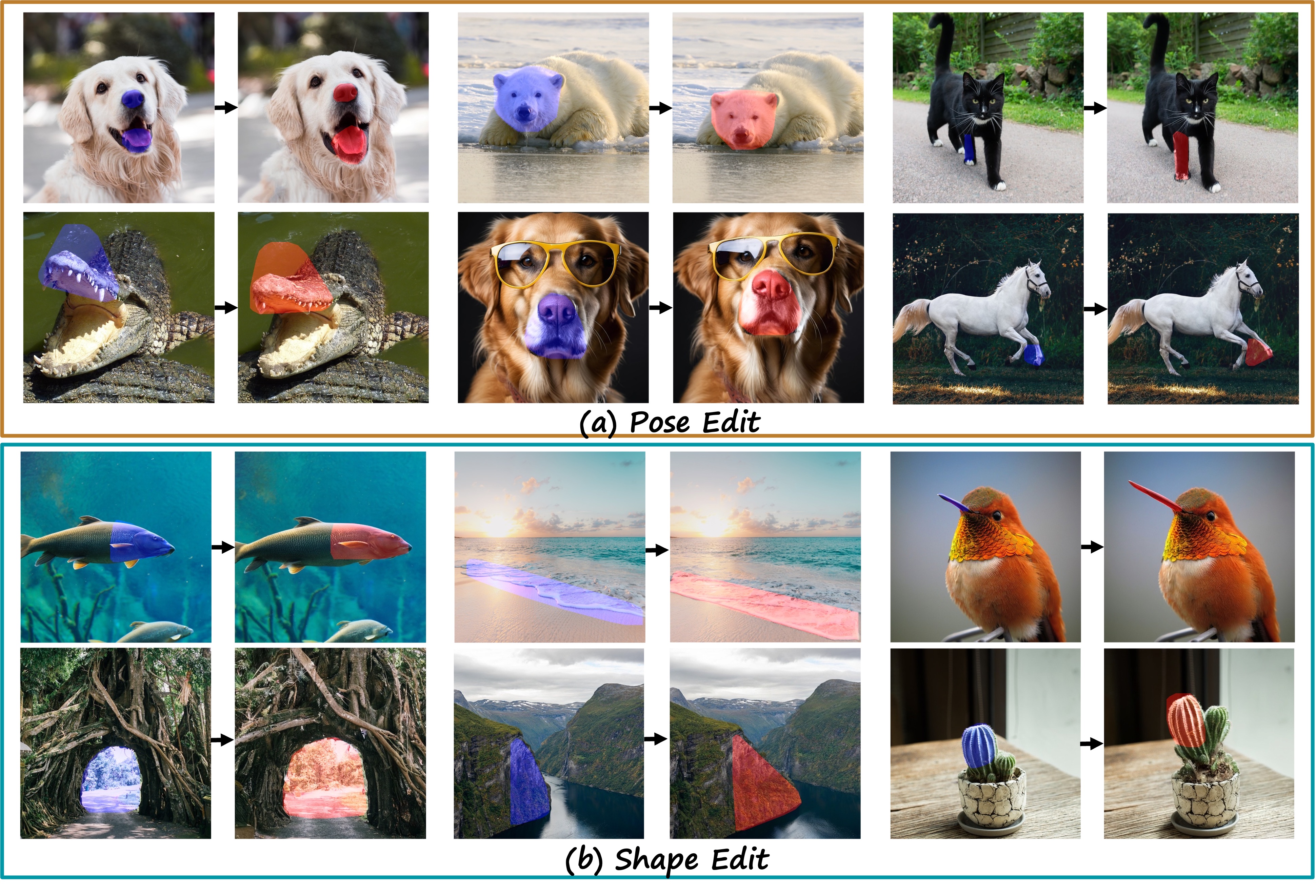}
    \captionof{figure}{Region-based Drag aims to transform the source region (blue mask) to align with the target region (red mask). Our In-Context Region-based Drag (ICRDrag) method supports fine-grained geometric editing like  pose or shape adjustment.}
    \label{fig:teaser}
\end{center}

\begin{abstract}

  Diffusion models have shown promise in drag-style editing. Previous works mainly focus on point-based drag, which is inherently ambiguous. This paper focuses on region-based drag and introduces a novel In-Context Region-based Drag (ICRDrag) method. Under the in-context learning framework, ICRDrag consumes a source image, a source region mask, and a target region mask, producing the target dragged image. Built upon the basic in-context learning model, we introduce two novel attention regularization: 1) image-mask attention consistency to ensure that a target region attends to similar source regions for image and mask modalities; 2) source-target attention correspondence to ensure the mutual correspondence between source and target regions. To facilitate region-based drag, we also construct Paired Region Dataset (PRD), a large-scale dataset with paired masks and images. Extensive experiments show that ICRDrag significantly outperforms existing methods in both quantitative metrics and user studies, achieving superior editing accuracy and visual fidelity. The dataset, code, and model are available at \url{https://github.com/bcmi/ICRDrag-Region-Drag-Editing}.
\end{abstract}

\section{Introduction}

Diffusion models\cite{sohl2015deep,ho2020denoising,kim2022diffusionclip,xu2023versatile,zhang2023adding,mou2023t2i,karnewar2023holodiffusion} have achieved remarkable success in diverse image generation and editing tasks, among which drag-style image editing \cite{pan2023drag,shi2023dragdiffusion,mou2024dragon,Mou2024DiffEditorBA,luo2023readout,nguyen2024edit,hou2024easydrag,liu2024drag,zhang2024gooddrag,combing,instantdrag,regiondrag,shi2024instadrag,avrahami2024diffuhaul,wang2025training,yan2025eedit,zhang2025framepainter,xia2024dreamomni,cai2024auto,choi2025dragtext,xia2025draglora,zhou2025dragnext,koo2025flowdrag,jiang2024clipdrag,chen2024adaptivedrag,pu2026dragging,yin2025lazydrag,liao2025directdrag,he2025contextdrag,yang2025attentiondrag,zhou2025dragflow} aims to drag partial regions in the image according to user-specified conditions. Based on the dragging condition, dragging task can be categorized into point-based drag and region-based drag.

In point-based drag, users provide pairs of source and target points. The source points in the image are expected to be dragged to the target points. However, this task suffers from inherent ambiguity. As discussed in RegionDrag\cite{regiondrag}, with limited point pairs, multiple plausible outcomes may exist, which are often misaligned with user intent. For instance, dragging a source point on a face towards a target point out of the face could mean either changing the facial orientation or stretching the face wider. Furthermore, due to the extreme sparsity of point-pair conditions, point-based methods have insufficient editing precision, that is, the source points can hardly be exactly dragged to the target points. Region-based drag \cite{ling2021editgan} addresses such ambiguity by using dense spatial conditions: users provide a source region mask (original location/shape) and a target region mask (desired location/shape), offering denser and more precise control that substantially reduces ambiguity.

Compared with point-based drag, there are very few works on region-based drag. EditGAN \cite{ling2021editgan} performs editing by optimizing the latent code to align with the target mask, but the interaction between image content and mask structure remains shallow — the mask primarily serves as a loss function rather than being deeply integrated into the generation process. RegionDrag \cite{regiondrag} adopts a copy-paste strategy in latent space by copying features from the source region and pasting them into the target region guided by the masks. This approach often leads to inconsistent boundaries where the pasted content does not seamlessly blend with the surrounding area, and it struggles with complex shape deformations that require more than simple feature transplantation.

The above limitations motivate us to explore a deeper integration of image and mask information into region-based drag. Recent advances in In-Context Learning (ICL) \cite{wang2023context,dong2022survey,huang2024context,selfsupericl,picl,Shi2023iclm} have shown that diffusion models exhibit inherent in-context capabilities, opening opportunities to treat structural cues like masks as conditioning context. Inspired by this, we propose In-context Region-based Drag (ICRDrag), built upon a DiT-based foundation model \cite{zhuo2024lumina}. ICRDrag takes a source image, a source mask, and a target mask as unified context to synthesize the target image in a single forward pass. 

Under this framework, we introduce two novel attention regularization. The first one is Image-Mask Attention Consistency (IMAC) regularization. We assume that a target region should attend to similar source regions for image and mask modalities, which enforces that the visual generation process is grounded on the spatial structures defined by the masks. The second one is Source-Target Attention Correspondence (STAC) regularization. We assume that the related regions in the source image and target image should mutually attend to each other, which reinforces the mutual correspondence between source and target regions. Additionally, we propose a novel two-stage training strategy that progressively increases task difficulty: the model first learns from complete region masks, then adapts to incomplete region masks containing only partial editing regions. This curriculum learning approach better simulates real-world sparse user inputs while ensuring stable training.

To the best of our knowledge, there is no large-scale dataset for region-based drag. Therefore, we construct Paired Region Dataset (PRD) from video dataset OpenVid \cite{nan2024openvid}, containing paired images and masks. Using SemanticSAM\cite{li2023semantic} and SAM2\cite{ravi2024sam2}, we extract multi-granularity segmentation masks with consistent labels and sample incomplete masks via optical flow. PRD training set contains 287,153 paired samples. Additionally, we construct PRDBench, a benchmark of 1,000 manually verified samples with both mask and point annotations.

We train our model on PRD, and evaluate on PRDBench and DragBench\cite{regiondrag}, comparing against both region-based and point-based methods. 
Extensive experiments show ICRDrag significantly outperforms existing methods in editing accuracy, visual realism, and detail preservation. Our contributions are three-fold: 1) We propose ICRDrag, an in-context learning framework for region-based drag; 2) We introduce two novel attention regularization and a novel curriculum training strategy; 3) We construct the PRD dataset to advance the research on region-based drag.

\section{Related Work}

\subsection{Point-based Dragging}
Existing point-based dragging methods can be categorized into two types based on their editing strategies. Some existing methods perform iterative, step-by-step editing to gradually transform the source point to the target point. Methods like DragGAN\cite{pan2023drag}, DragDiffusion\cite{shi2023dragdiffusion}, 
SDE-Drag\cite{nie2023blessing},
FreeDrag \cite{ling2023freedrag}, CLIPDrag\cite{jiang2024clipdrag}, StableDrag\cite{cui2024stabledrag}, EasyDrag\cite{hou2024easydrag}, DragNoise \cite{liu2024drag}, GoodDrag \cite{zhang2024gooddrag}, AdaptiveDrag \cite{chen2024adaptivedrag}, FlowDrag\cite{koo2025flowdrag}, DirectDrag\cite{liao2025directdrag} and DragLoRA\cite{xia2025draglora} fall into this type. DragGAN pioneers motion supervision and point tracking using StyleGAN, while DragDiffusion introduces diffusion models into this task. Subsequent works improve point tracking (e.g., EasyDrag, StableDrag), optimization strategies (e.g., DragNoise, GoodDrag), or enhance semantic understanding (e.g., AdaptiveDrag). Other approaches such as DragonDiffusion \cite{mou2024dragon}, DragAPart \cite{li2024dragapart}, FastDrag \cite{zhao2024fastdrag}, LightningDrag \cite{shi2024instadrag}, LucidDrag \cite{cui2024localize}, InstantDrag \cite{instantdrag}, GeoDrag\cite{liu2024drag}, AttentionDrag\cite{yang2025attentiondrag}, LazyDrag\cite{yin2025lazydrag}, ContextDrag\cite{he2025contextdrag} and Inpaint4Drag\cite{lu2025inpaint4drag} edit images in a single forward pass. These methods incorporate innovations like classifier guidance, latent warping functions, large vision-language models, and optical flow prediction for conditioning the editing process.

\subsection{Region-based Dragging}

EditGAN \cite{ling2021editgan} initiates an interactive paradigm of region-based drag, in which both source and target region masks are employed as control signals. The source region mask indicates the original shape or position of the object to be edited, while the target region mask specifies the desired shape or position after editing. Similar to EditGAN \cite{ling2021editgan}, Pixel-Guided Diffusion \cite{matsunaga2022fine} inherits the interactive setting. Both Pixel-Guided Diffusion and EditGAN perform editing by predicting segmentation masks and optimizing the latent code to align with the target region mask. Differently, RegionDrag \cite{regiondrag} adopts a copy-and-paste strategy in the latent space to achieve region-based dragging. Notably, RegionDrag draws inspiration from point-based dragging methods. It first transforms the editing regions into a set of point pairs, and then performs the latent copy-and-paste operation accordingly. More recently, DragFlow \cite{zhou2025dragflow} proposes a training-free method built upon Diffusion Transformer (DiT) architecture,  which addresses the poor performance of conventional point-based supervision on fine-grained DiT features by introducing region-level affine supervision.

\section{Problem Definition}
\label{formulation}

The user provides a source image $\bm{I}_s$ and the corresponding source region mask $\bm{M}_s$ that specifies the regions to be edited. In this mask, users can specify multiple editing regions, where each one is assigned with a unique label ID. This mask can be obtained using segmentation models such as SAM \cite{kirillov2023segment}, or drawn by users. The user also provides a target region mask $\bm{M}_g$, which reflects the modifications to the source mask. In this mask, the label IDs of the editing regions should remain consistent with those in the source region mask. The goal of region-based drag is to generate a target edited image $\bm{I}_e$ from source image $\bm{I}_s$, based on the transformation defined by source/target region masks.  $\bm{I}_e$ is expected to approach the ground-truth target image $\bm{I}_g$.

For notational convenience, we use ${\hat{\bm{I}}_s, \hat{\bm{M}}_s, \hat{\bm{I}}_g, \hat{\bm{M}}_g} \in \mathbb{R}^{H \times W \times D}$ to denote the latent codes of ${\bm{I}_s, \bm{M}_s, \bm{I}_g, \bm{M}_g}$, in which $H$, $W$, $D$ are the height, width, channel number of latent code respectively. The objective is to learn a mapping function $\mathcal{F}$ from $\{\hat{\bm{I}}_s, \hat{\bm{M}}_s, \hat{\bm{M}}_g\}$ to $\hat{\bm{I}}_g$:
\begin{eqnarray} \label{eqn:mapping}
\mathcal{F}_{\bm{\theta}}(\hat{\bm{I}}_s, \hat{\bm{M}}_s, \hat{\bm{M}}_g): \mathbb{R}^{3 \times H \times W \times D} \rightarrow \mathbb{R}^{H \times W \times D},
\end{eqnarray}
where $\bm{\theta}$ represents the model parameters.

\begin{figure*}[t]
  \centering
  \includegraphics[width=.8\linewidth]{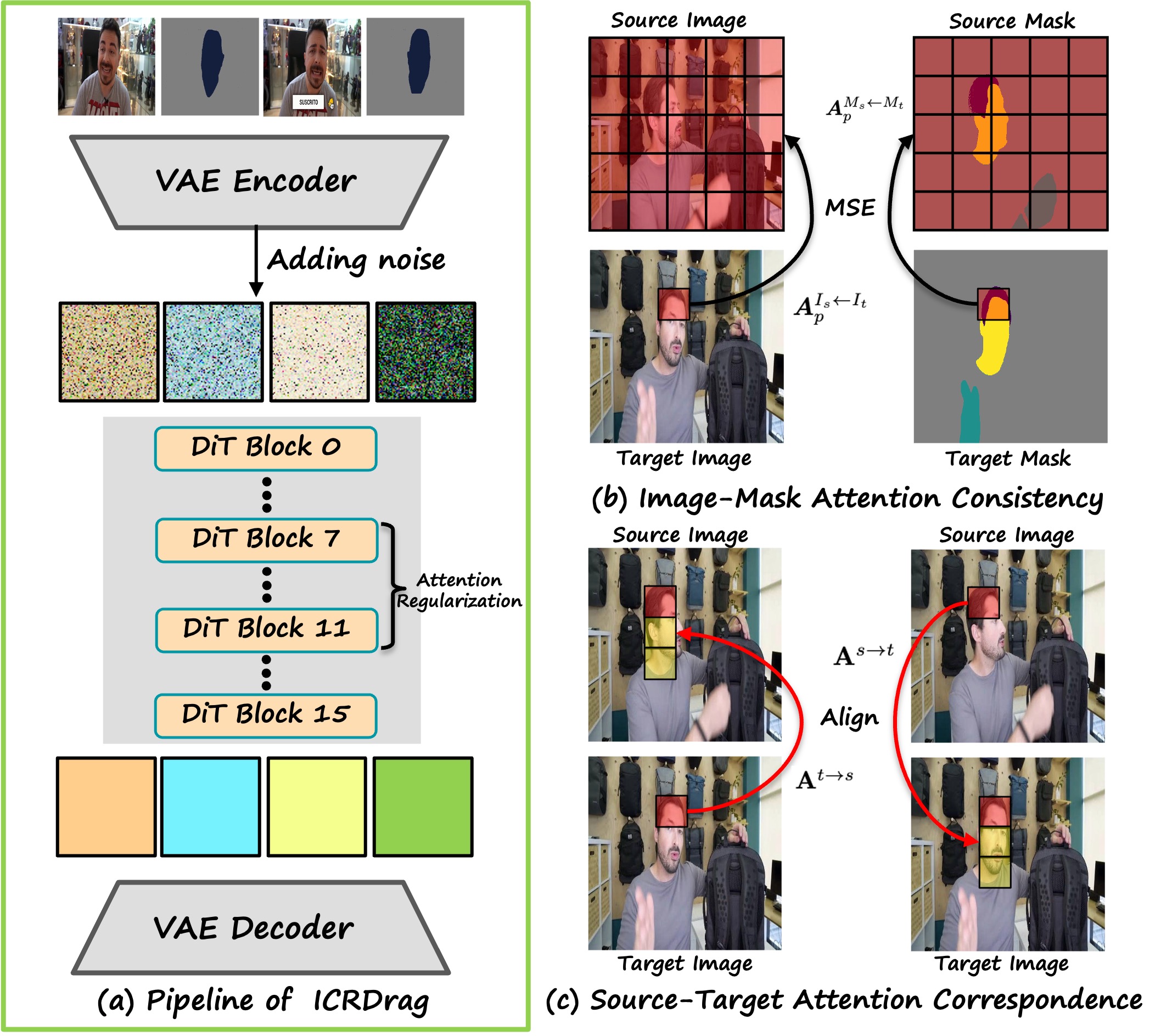}
  \caption{ (a) The overall pipeline of ICRDrag. (b) Image-Mask Attention Consistency. For one patch in the target image, its attention over the source image should mirror the attention of the corresponding patch in the target mask over the source mask. (c) Source-Target Attention Correspondence. If a target patch attends to a source patch, that source patch should also attend back to the same target patch.}
  \label{fig:overall}
\end{figure*}

\section{ICRDrag}
\label{method}

In Section \ref{overall}, we will introduce the in-context learning framework for region-based drag. In Section \ref{sec:image-mask} and Section \ref{sec:source-target}, we will present two novel attention regularization: image-mask attention consistency and source-target attention correspondence. 
In Section \ref{twostage}, we will introduce our two-stage curriculum training strategy.

\subsection{Overall Architecture}
\label{overall}

Our proposed method ICRDrag is model-agnostic and applicable across different DiT architectures. As shown in Figure \ref{fig:overall}(a), we build our model upon Next-DiT\cite{zhuo2024lumina} due to its strong generative performance across multiple modalities, and adapt it to region-based drag. The latent codes $\hat{\bm{I}}_s, \hat{\bm{M}}_s, \hat{\bm{M}}_g$ obtained via VAE encoder serve as conditional inputs and remain noise-free throughout both training and inference. The target image latent $\hat{\bm{I}}_g$ is the only variable that undergoes noising and denoising. The model predicts a velocity field, which is used to denoise the target latent, and the final denoised latent is passed through the VAE decoder to synthesize the edited image.

\paragraph{Modality-specific LoRAs.}  Images and masks have intrinsically different properties: images are rich in texture and fine-grained details, while masks are sparse and encode only spatial structure information. Processing them through shared parameters risks feature confusion, that is, the mask's structural representations may contaminate the image pathway, leading to the loss of details and overly smooth generations. To address this issue, we incorporate separate LoRA \cite{hu2022lora} modules into the feed-forward networks: one for image tokens and another for mask tokens. This allows each modality to learn representations suited to its own properties without cross-modality interference. 

\paragraph{Training.} At each training iteration, we sample a timestep $t \sim \text{LogNorm}(0, 1)$ \cite{esser2024scaling}, along with Gaussian noise $\bm{\epsilon} \sim \mathcal{N}(0, I)$. The conditional inputs $\{\hat{\bm{I}}_s, \hat{\bm{M}}_s, \hat{\bm{M}}_g\}$ remain in their noise-free state. Only the target image $\hat{\bm{I}}_g$ is corrupted by adding noise:
\begin{equation}
    \hat{\bm{I}}_g^{t} = (1-t) \hat{\bm{I}}_g + t\bm{\epsilon}.
    \label{eq:noising}
\end{equation}

The velocity field for the target image is defined as $\bm{u} = \hat{\bm{I}}_g - \bm{\epsilon}$. The training objective is pushing the predicted velocity field towards the target velocity $\bm{u}$ via the flow-matching loss:
\begin{equation}
    \mathcal{L}_{\text{flow}} = \mathbb{E} \left[ \left\| \bm{v_\theta}(t, \hat{\bm{I}}_s, \hat{\bm{M}}_s, \hat{\bm{I}}_g^{t}, \hat{\bm{M}}_g) - \bm{u} \right\|^2 \right],
    \label{eq:loss}
\end{equation}
where $\bm{\theta}$ represents the model parameters.

\paragraph{Inference.} During inference, our goal is to generate the target edited image $\bm{I}_e$ conditioned on $\{\bm{I}_s, \bm{M}_s, \bm{M}_g\}$. We initialize the latent code of the target image $\hat{\bm{I}}_e^T$ with random Gaussian noise at the starting timestep $T$. The conditional inputs remain noise-free throughout the denoising process. At each denoising step $t$ (from $T$ down to $0$), the model predicts the velocity field to update $\hat{\bm{I}}_e^t$ towards the clean target image. After completing all denoising steps, the final latent $\hat{\bm{I}}_e^0$ is passed through the VAE decoder to obtain the edited image $\bm{I}_e$.

\subsection{Image-Mask Attention Consistency}
\label{sec:image-mask}

We assume that a target region should attend to similar source regions for image
and mask modalities, leading to Image-Mask Attention Consistency (IMAC) regularization. Specifically, for each patch in the target image, its attention over the source image to gather visual features should mirror the attention of the corresponding patch in the target mask over the source mask to understand the spatial structures. Such alignment ensures that the visual generation process is grounded on the spatial structures defined by the masks. 

As shown in Figure \ref{fig:overall}(b), the source and target masks contain the spatial structural information for hair, face, and arm. During model training, it could be easily learned that the hair (\emph{resp.}, face, arm) patch in the target mask should attend to the hair (\emph{resp.}, face, arm) patch in the source mask. By aligning image attention with mask attention, the generation of hair (\emph{resp.}, face, arm) patch in the target image could better gather the visual features from the hair (\emph{resp.}, face, arm) patch in the source image. 

Formally, let $\mathcal{P}_g$ denote the set of patches in the target image that lie within the target region mask $\bm{M}_g$. For each patch $p \in \mathcal{P}_g$, we extract its corresponding token from the target image branch and use it as a query to compute attention over all patches in the source image, yielding the attention map $\bm{A}_p^{I_s \leftarrow I_g} \in \mathbb{R}^{N_s}$, where $N_s$ is the number of patches in the source image. Similarly, we take the token of the corresponding patch in the target mask branch (\emph{i.e.}, the patch at the same spatial location $p$) as a query to compute attention over all patches in the source mask, yielding the attention map $\bm{A}_p^{M_s \leftarrow M_g} \in \mathbb{R}^{N_s}$. We enforce consistency between these two attention maps by minimizing their discrepancy:
\begin{equation} 
\mathcal{L}_{\text{imac}} = \sum_{p \in \mathcal{P}_g} \left\| \bm{A}_p^{I_s \leftarrow I_g} - \bm{A}_p^{M_s \leftarrow M_g} \right\|^2.
\label{eq:img_mask}
\end{equation}

\subsection{Source-Target Attention Correspondence}
\label{sec:source-target}

We assume that the related regions in the source image and target image should mutually attend to each other, leading to the Source-Target Attention Correspondence (STAC) regularization. As shown in Figure \ref{fig:overall}(c), if a target patch attends to a source patch, then that source patch should also attend back to the same target patch. Such mutual reinforcement ensures that the model establishes consistent correspondences between the source and target images, which is essential for accurate spatial transformations such as object movement, resizing, and articulation.

Formally, let $\mathcal{P}_g$ be the set of target image patches within the target region mask, and $\mathcal{P}_s$ be the set of source image patches within the source region mask. We consider two attention matrices $\mathbf{A}^{g \rightarrow s}$ and $\mathbf{A}^{s \rightarrow g}$. $\mathbf{A}^{g \rightarrow s} \in \mathbb{R}^{|\mathcal{P}_g| \times |\mathcal{P}_s|}$ denotes the attention from target patches to source patches, where each entry $\mathbf{A}^{g \rightarrow s}_{ij}$ is the attention value from target patch $i$ (query) to source patch $j$. $\mathbf{A}^{s \rightarrow g} \in \mathbb{R}^{|\mathcal{P}_s| \times |\mathcal{P}_g|}$ denotes the attention from source patches to target patches, where each entry $\mathbf{A}^{s \rightarrow g}_{ji}$ means the attention value from source patch $j$ (query) to target patch $i$.

We assume that if target patch $i$ strongly attends to source patch $j$, then source patch $j$ should also strongly attend to target patch $i$. Such mutual correspondence can be captured by the diagonal entries in the product $\mathbf{A}^{g \rightarrow s} \mathbf{A}^{s \rightarrow g} \in \mathbb{R}^{|\mathcal{P}_g| \times |\mathcal{P}_g|}$. In particular, for the target patch $i$, the diagonal entry \allowbreak $(\mathbf{A}^{g \rightarrow s} \mathbf{A}^{s \rightarrow g})_{ii}$ $=$ $\sum_{j} \mathbf{A}^{g \rightarrow s}_{ij} \mathbf{A}^{s \rightarrow g}_{ji}$ sums up the product of its mutual attentions over all source patches. When maximizing $(\mathbf{A}^{g \rightarrow s} \mathbf{A}^{s \rightarrow g})_{ii}$,  the source patches that the target patch $i$ attends to are enforced to attend back to the target patch $i$. Therefore, we aim to minimize the following loss function:
\begin{equation} 
\mathcal{L}_{\text{stac}} = -\operatorname{Trace}\left( \mathbf{A}^{g \rightarrow s} \mathbf{A}^{s \rightarrow g} \right),
\label{eq:src_tgt}
\end{equation}
where $\operatorname{Trace}(\cdot)$ denotes the trace operator. Minimizing the negative trace encourages the model to maximize the diagonal entries, thereby enforcing mutual correspondence between source and target patches.

Besides the basic flow-matching loss $\mathcal{L}_{\text{flow}}$ in Eqn.~\ref{eq:loss}, we add two auxiliary losses $\mathcal{L}_{\text{imac}}$ in Eqn.~\ref{eq:img_mask} and $\mathcal{L}_{\text{stac}}$ in Eqn.~\ref{eq:src_tgt}, leading to the following total loss:
\begin{equation}
\mathcal{L}_{\text{total}} = \mathcal{L}_{\text{flow}} + \lambda_1 \mathcal{L}_{\text{imac}} + \lambda_2 \mathcal{L}_{\text{stac}},
\label{eq:total_loss}
\end{equation}
where $\lambda_1$ and $\lambda_2$ are hyper-parameters.  

Both auxiliary losses are computed using attention maps extracted from the $7^{th}-11^{th}$  layers of the transformer (we elaborate on the rationale for selecting these layers to design the losses in Section \ref{sec:attention_analysis}), normalized via softmax. For $\mathcal{L}_{\text{imac}}$, we directly compute MSE on two attention maps. For $\mathcal{L}_{\text{stac}}$, we average attention across multiple heads before computing the trace of the product of bidirectional attention matrices.

\subsection{Two-stage Curriculum Training Strategy}
\label{twostage}
In real-world applications, the source and target region masks may be very sparse, in which only a few regions are separated out. We refer to the region masks with all semantic regions separated out as complete region mask, and the region masks with partial regions separated out as incomplete region mask (see Section \ref{sec:datacon} for detailed illustration). 

Compared with complete region mask, dragging based on incomplete region masks poses a greater challenge for the model, due to the following reasons. Incomplete region masks contain much sparser and less information than complete region masks. In some scenarios, drag-style editing may imply changes beyond the explicitly separated regions. Therefore, the model must also learn to infer the necessary adjustments in those unseparated regions.

Following the routine of curriculum learning \cite{hacohen2019power,pentina2015curriculum,wang2021survey,bengio2009curriculum}, we design a training strategy that progressively increases task difficulty, guiding the model to learn from easy to hard setting. To reduce the difficulty of model training and ensure smoother optimization process, we design a two-stage training strategy.

\noindent{\textbf{Training stage 1}}: We obtain the latent codes of source image $\hat{\bm{I}}_s$, target image $\hat{\bm{I}}_g$, source complete region mask $\hat{\bm{M}}_s'$, and target complete region mask $\hat{\bm{M}}_g'$ via VAE encoder. The conditional inputs ($\hat{\bm{I}}_s$, $\hat{\bm{M}}_s'$, $\hat{\bm{M}}_g'$) remain noise-free throughout training. Only the target image latent $\hat{\bm{I}}_g$ is corrupted by adding noise at a randomly sampled timestep $t$. 

\noindent{\textbf{Training stage 2}}: This stage is similar to Training Stage 1 and the key difference lies in region masks. Instead of using complete region masks as input, we construct incomplete region masks $\bm{M}_s, \bm{M}_g$ by sampling specific regions from the complete $\bm{M}_s', \bm{M}_g'$. We only retain the sampled regions while filling the remaining regions with gray value. To enhance the model robustness to potentially inaccurate user-provided region masks, we randomly apply dilation to the sampled regions.
\begin{figure*}[t!]
  \centering
  \includegraphics[width=.8\textwidth]{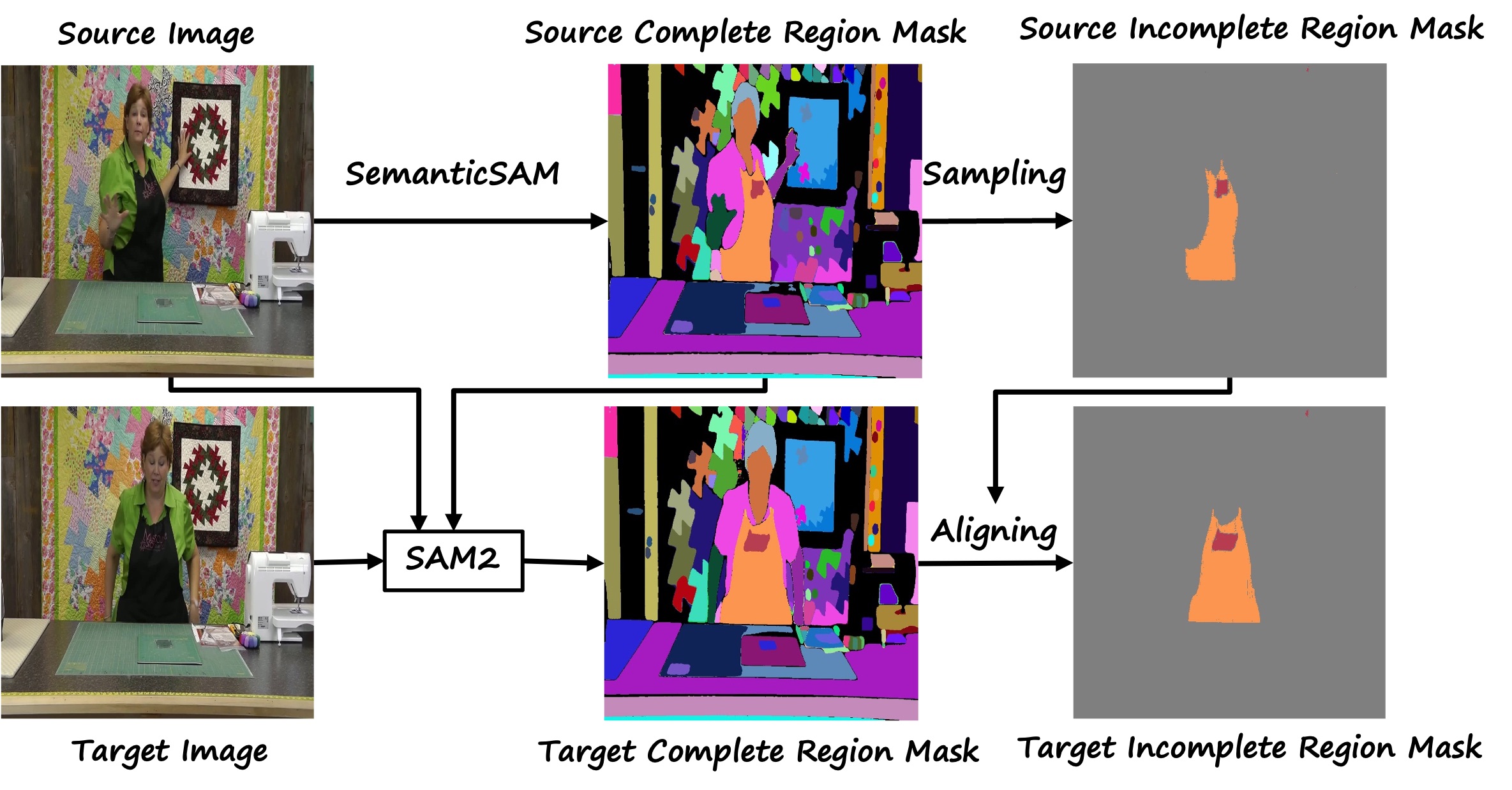}
  \caption{Paired Region Dataset construction. We leverage SemanticSAM \cite{li2023semantic} and SAM2 \cite{ravi2024sam2} to generate fine-grained segmentation masks. Incomplete region masks are then sampled based on estimated optical flow  combined with the watershed algorithm.}
  \label{fig:datacon}
\end{figure*}
\section{Paired Region Dataset (PRD)}
\label{sec:datacon}

To the best of our knowledge, there is no existing large-scale dataset tailored to region-based drag. Therefore, we construct the Paired Region Dataset (PRD) consisting of two components: a large-scale training set for model training, and a high-quality benchmark for model evaluation.

\paragraph{Training set construction.}
We construct the training set based on the OpenVid \cite{nan2024openvid} video dataset, which contains one million high-quality video clips accompanied by expressive captions. 

Considering that users may edit an image at different levels of granularity (\emph{e.g.}, coarse-grained face mask as a whole versus fine-grained part masks including ears, nose, and eyes), it is necessary to obtain multi-granularity segmentation masks while ensuring consistency in terms of segmentation labels, that is, the same region should have the same segmentation label across the source and target images. To achieve this goal, we use SemanticSAM \cite{li2023semantic} to extract multi-granularity segmentation masks from the source image, and then feed both the segmentation mask and the target image into SAM2 \cite{ravi2024sam2} to obtain the corresponding segmentation mask for the target image. In this way, we acquire source images $\bm{I}_s$, source complete region masks $\bm{M}_s'$, target images $\bm{I}_g$, and target complete region masks $\bm{M}_g'$.

To fulfill the requirements of the second-stage training, we need to obtain incomplete region masks. We first compute the optical flow between the source image $\bm{I}_s$ and the target image $\bm{I}_g$ using UniMatch \cite{xu2023unifying}. Based on the estimated optical flow, we apply the watershed algorithm to sample 1–5 keypoints as source points. The corresponding target points are then obtained by adding the optical flow vectors to the source points. We extract the regions in the source complete region mask $\bm{M}_s'$ containing the start points, and those in the target complete region mask $\bm{M}_g'$ containing the end points, leading to incomplete region masks $\bm{M}_s, \bm{M}_g$.

In total, we obtain 287,153 tuples of source image, source region mask, target image, target region mask for the training set.

\paragraph{Benchmark construction.}
Besides the large-scale training set, a high-quality benchmark is essential for model evaluation. Therefore, we construct a benchmark of 1,000 manually refined and annotated samples through the following procedure.

\noindent \textbf{Data selection:} We apply the same pipeline as constructing the training set to the remaining raw data that are not included in the training split. From the processed candidate pairs, we select samples to ensure diversity in object categories, editing types (\emph{e.g.}, position change, resizing, shape deformation, pose adjustment), and complexity levels.

\noindent \textbf{Verification and correction:} For each candidate sample, annotators check the validity of each dragging scenario, including stable viewpoint, no object insertion or removal, and reasonable transformation magnitude. They further verify the consistency of source/target points and masks with the intended edit. Samples that do not meet these criteria are discarded or re-annotated.

\noindent \textbf{Keypoint annotation and mask derivation:} For samples that require refinement, annotators re-establish the editing correspondences by selecting a set of source points $\{q_s^k|_{k=1}^{n}\}$ and corresponding target points $\{q_g^k|_{k=1}^{n}\}$ . Each point pair is constrained to share the same segmentation label to maintain semantic consistency. The number of point pairs $n$ is in the range of $[1, 5]$. This design facilitates a fair comparison between point-based and region-based dragging approaches. Region masks are then automatically derived by extracting the regions containing the annotated points from the complete region masks $\bm{M}_s'$ and $\bm{M}_g'$, forming the incomplete region masks $\bm{M}_s$ and $\bm{M}_g$.

The final benchmark consists of 1,000 high-quality tuples, each containing: (1) source image $\bm{I}_s$ and target image $\bm{I}_g$; (2) source region mask $\bm{M}_s$ and target region mask $\bm{M}_g$; (3) A set of source points $\{q_s^k|_{k=1}^{n}\}$  and corresponding target points $\{q_g^k|_{k=1}^{n}\}$.

\begin{figure*}[t!]
  \centering
  \includegraphics[width=.95\textwidth]{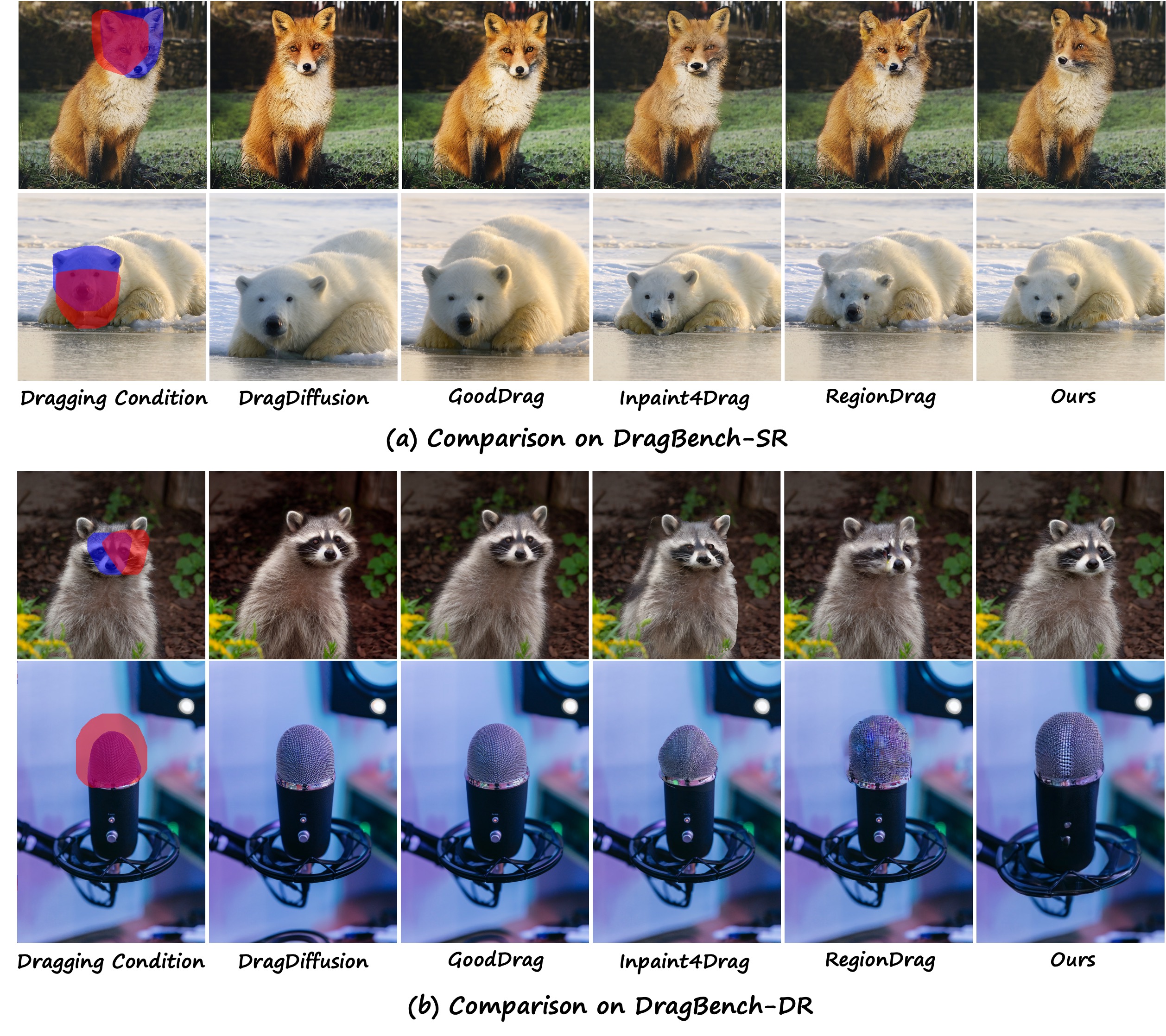}
  \caption{Qualitative results on DragBench-SR and DragBench-DR \cite{regiondrag}. In the “Dragging Condition” column, the blue mask indicates the source region, while the red mask indicates the target region.}
  \label{fig:comparison_on_dragbench}
\end{figure*}

\begin{figure*}
  \centering
  \includegraphics[width=.9\textwidth]{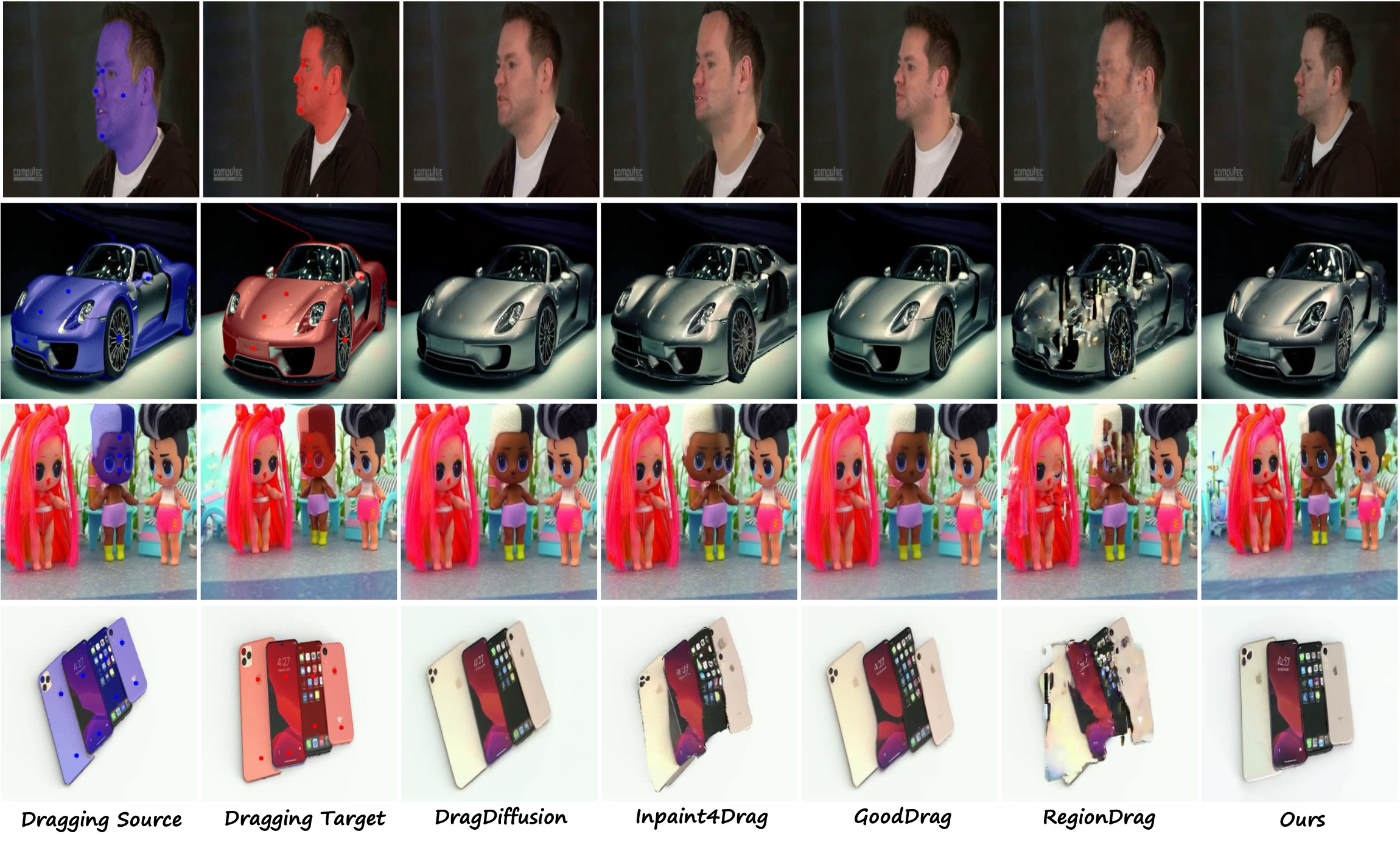}
  \caption{Visual results on our PRD benchmark.
  }
  \label{fig:viz_on_openvid}
\end{figure*}

\begin{table*}[t]
  \centering
  \caption{Quantitative analysis on PRD benchmark.}
  \label{tab:comparison}
  \begin{tabular}{c|ccccc}
    \toprule
       Method         & MSE $\downarrow$    & LPIPS$\downarrow$ & SSIM $\uparrow$& MD(RegionDrag)$\downarrow$ & MD(DragLoRA)$\downarrow$\\
    \midrule
    DragDiffusion\cite{shi2023dragdiffusion} & 0.0937 & 0.1836 & 0.5993 & 5.17 & 26.79 \\
    SDE-Drag\cite{nie2023blessing} & 0.1017 & 0.2018 & 0.5849 & 7.97 & 44.04\\
    DiffEditor\cite{Mou2024DiffEditorBA} & 0.0959 & 0.1949 & 0.6071 & 23.45 & 31.19 \\
    FastDrag\cite{zhao2024fastdrag} & 0.0962 & 0.2049 & 0.5862 & 6.01 & 31.22\\
    Inpaint4Drag\cite{lu2025inpaint4drag} & 0.0972 & 0.1923 & \underline{0.6102} & 4.24 & 23.62\\
    GoodDrag\cite{zhang2024gooddrag} & \underline{0.0902} & \underline{0.1761} & 0.6094 & \underline{3.70} & \textbf{18.01} \\
    DragLoRA\cite{xia2025draglora} & 0.0933 & 0.1870 & 0.5974 & 4.62 & 23.96 \\
    \midrule
    RegionDrag\cite{regiondrag} & 0.0977 & 0.1944 & 0.6076 & 8.02 & 43.05 \\
    ICRDrag   & \textbf{0.0735} & \textbf{0.1610} & \textbf{0.6284} & \textbf{3.66} & \underline{22.34} \\
    \bottomrule
  \end{tabular}
\end{table*}

\begin{table*}[t]
  \centering
  \caption{User study results on DragBench.}
  \label{tab:userstudy}
  \begin{tabular}{c|ccc}
    \toprule
     Method           & Realism$\uparrow$ & Fidelity$\uparrow$ & Region Accuracy$\uparrow$\\
    \midrule
    GoodDrag\cite{zhang2024gooddrag} & 0.2876 & 0.2120 & 0.1276\\
    RegionDrag\cite{regiondrag} & 0.2442 & 0.2314 & 0.3518\\
    ICRDrag & \textbf{0.4682} & \textbf{0.5566} &\textbf{0.5206} \\
    \bottomrule
  \end{tabular}
\end{table*}

\section{Experiment}
\label{sec:exp}

\subsection{Experimental Setting}
\label{sec:exp_setting}

\noindent{\textbf{Implementation details.}} 
During the training stage 1, we train the model for 60,000 steps with batch size 2 and learning rate $1 \times 10^{-4}$. While for training stage 2, we train the model for another 2,000 steps, with batch size 1 and learning rate $5 \times 10^{-5}$. More implementation details are left to supplementary.

\noindent{\textbf{Baseline.}} We compare our method against both region-based and point-based dragging approaches.
For region-based drag, we compare with RegionDrag \cite{regiondrag}. While EditGAN \cite{ling2021editgan} and Pixel-Guided Diffusion \cite{matsunaga2022fine} also fall into this category, they are less suitable for general-domain evaluation. EditGAN is built upon GAN architectures, and Pixel-Guided Diffusion relies on DDPMSegmentation\cite{baranchuk2021labelefficient}. Their design choices limit their applicability to general-domain images. Therefore, we focus our region-based comparison with those methods designed for general-purpose editing.
For point-based dragging, we select representative methods that cover diverse technical approaches: DragDiffusion \cite{shi2023dragdiffusion} as a pioneering diffusion-based method, SDE-Drag\cite{nie2023blessing}, GoodDrag\cite{zhang2024gooddrag} and DragLoRA\cite{xia2025draglora} for its optimization-based improvements, and DiffEditor\cite{Mou2024DiffEditorBA}, FastDrag\cite{zhao2024fastdrag}, Inpaint4Drag\cite{lu2025inpaint4drag}, as recent one-step editing approaches. 
For training-based baseline DiffEditor\cite{Mou2024DiffEditorBA}, we fine-tune it on our PRD training set to ensure fair comparison.

\noindent{\textbf{Dataset and metrics.}} 
We train our model on our Paired Region Dataset (PRD). We evaluate our proposed method on PRD benchmark and DragBench~\cite{regiondrag}, which includes DragBench-DR and DragBench-SR. Note that PRD benchmark has ground-truth images while the other two do not have.  On PRD benchmark, we adopt LPIPS~\cite{zhang2018unreasonable}, SSIM, MSE, and Mean Distance (MD) to measure the difference between editing result and ground-truth. On DragBench-DR and DragBench-SR, we conduct user study from three aspects: realism, fidelity, and region accuracy.  

\subsection{Experimental Result}

\textbf{Quantitative analysis.}
We quantitatively compare our method with existing drag-style editing methods on PRD benchmark. As shown in Table \ref{tab:comparison}, our method consistently outperforms existing methods across most metrics. It achieves lower LPIPS and higher SSIM scores, indicating better visual fidelity and detail preservation. Additionally, lower MSE and lower MD indicate more accurate and controllable editing.

\noindent{\textbf{Qualitative analysis.}}
Figure \ref{fig:comparison_on_dragbench} compares our method ICRDrag with point-based and region-based methods on DragBench. ICRDrag achieves superior performance in editing accuracy, realism, artifact reduction, and global consistency. On the more challenging PRD dataset (Figure \ref{fig:viz_on_openvid}), ICRDrag demonstrates robust results across complex cases. In the third example of Figure \ref{fig:viz_on_openvid}, point-based methods introduce ambiguity by dragging the person toward the left side of the image, while region-based drag better aligns with the editing objective. Moreover, GoodDrag, a representative point-based method, exhibits noticeable degradation in editing fidelity and realism in complex scenes. In contrast, our ICRDrag consistently maintains high accuracy and visual quality. Additional hard-case examples and cross-dataset transfer results are provided in supplementary.

\noindent{\textbf{User study.}}
We obtain images from DragBench and present original image, editing region masks, and the edited results from GoodDrag, RegionDrag and ICRDrag to 50 users. Participants were asked to choose the best result in terms of realism, fidelity, and region accuracy. The percentage of preferred results is summarized in Table \ref{tab:userstudy}. Since GoodDrag can only take point pairs as input, its region accuracy is much worse than other methods.

\begin{figure*}[t!]
  \centering
  \includegraphics[width=0.9\textwidth]{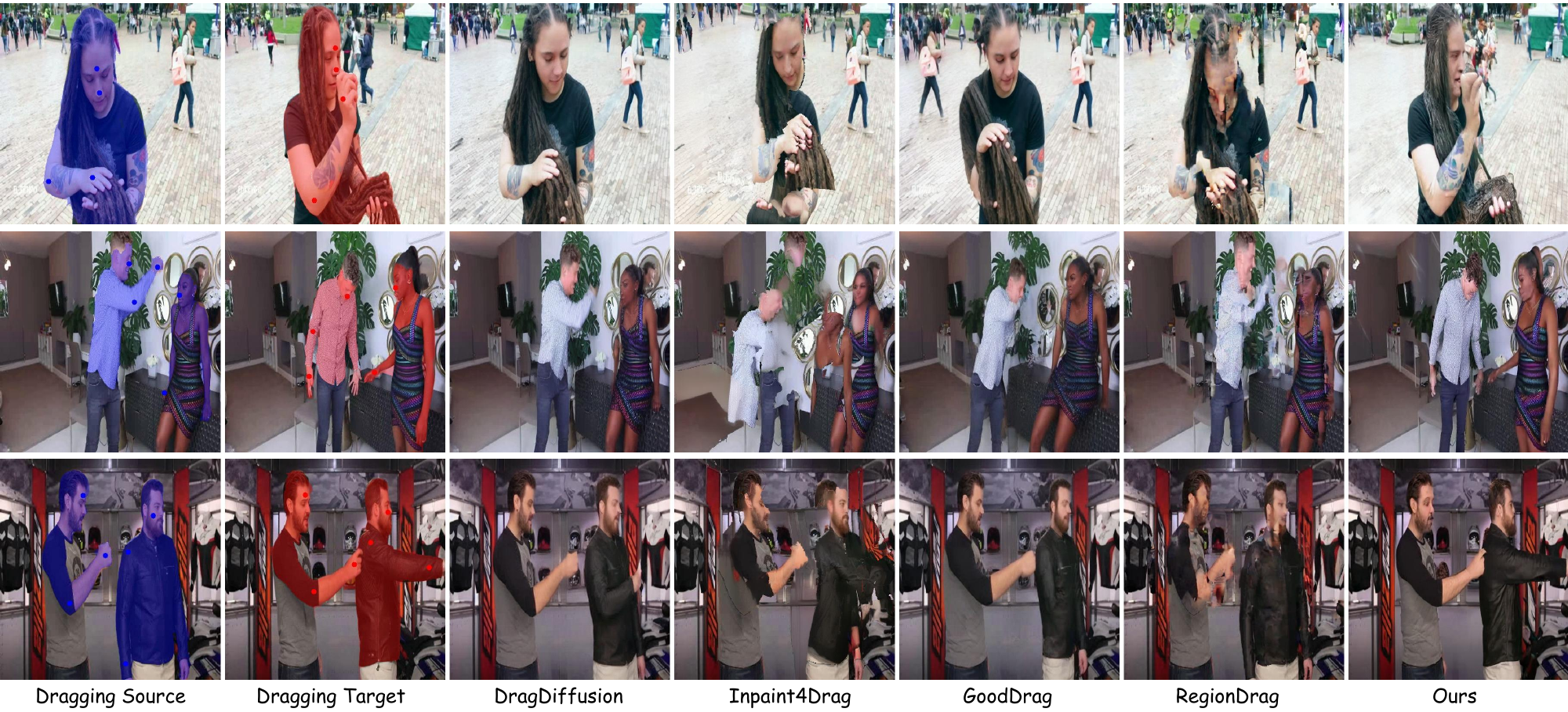}
  \caption{Comparison with baselines on hard cases involving large topology changes, occlusion, and human limb repositioning.}
  \label{fig:hardcase}
\end{figure*}

\begin{figure*}[t!]
  \centering
  \includegraphics[width=0.9\textwidth]{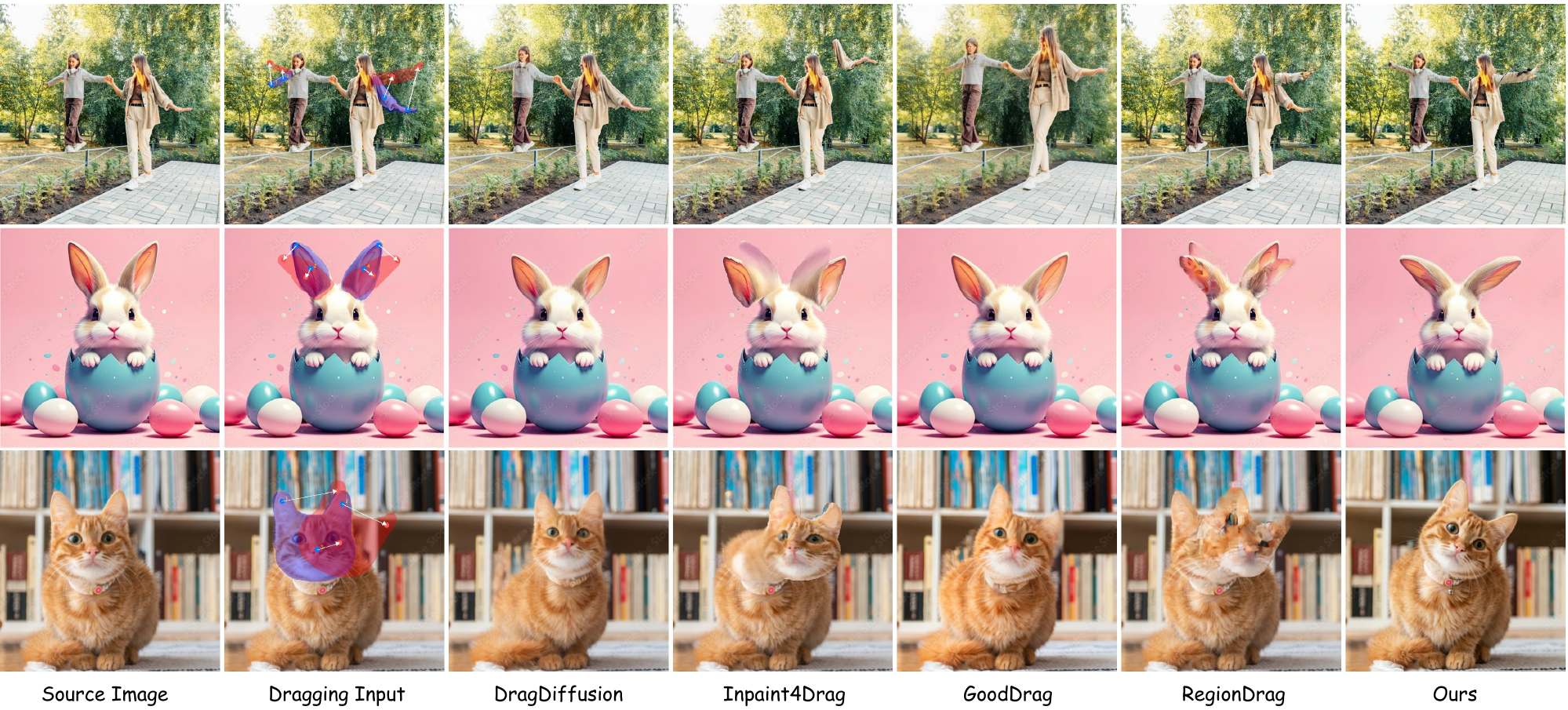}
  \caption{Comparison with baselines on cross-dataset transfer examples collected from Adobe Stock.}
  \label{fig:adobecase}
\end{figure*}

\subsection{Hard Cases and Cross-dataset Transfer}
\label{sec:hard_transfer}

We further provide qualitative comparisons on challenging non-rigid editing scenarios and out-of-distribution images. These examples complement results by covering cases that go beyond simple translation or scale changes. As shown in Figure~\ref{fig:hardcase}, ICRDrag handles hard cases involving large topology changes, partial occlusion, and human limb repositioning. Figure~\ref{fig:adobecase} presents cross-dataset transfer results on images collected from Adobe Stock. Without changing the inference setting, ICRDrag still follows the source-target region masks and produces plausible edits on these real-world images, indicating its generalization ability beyond PRD and DragBench.

\begin{figure*}[t!]
  \centering
  \includegraphics[width=.75\textwidth]{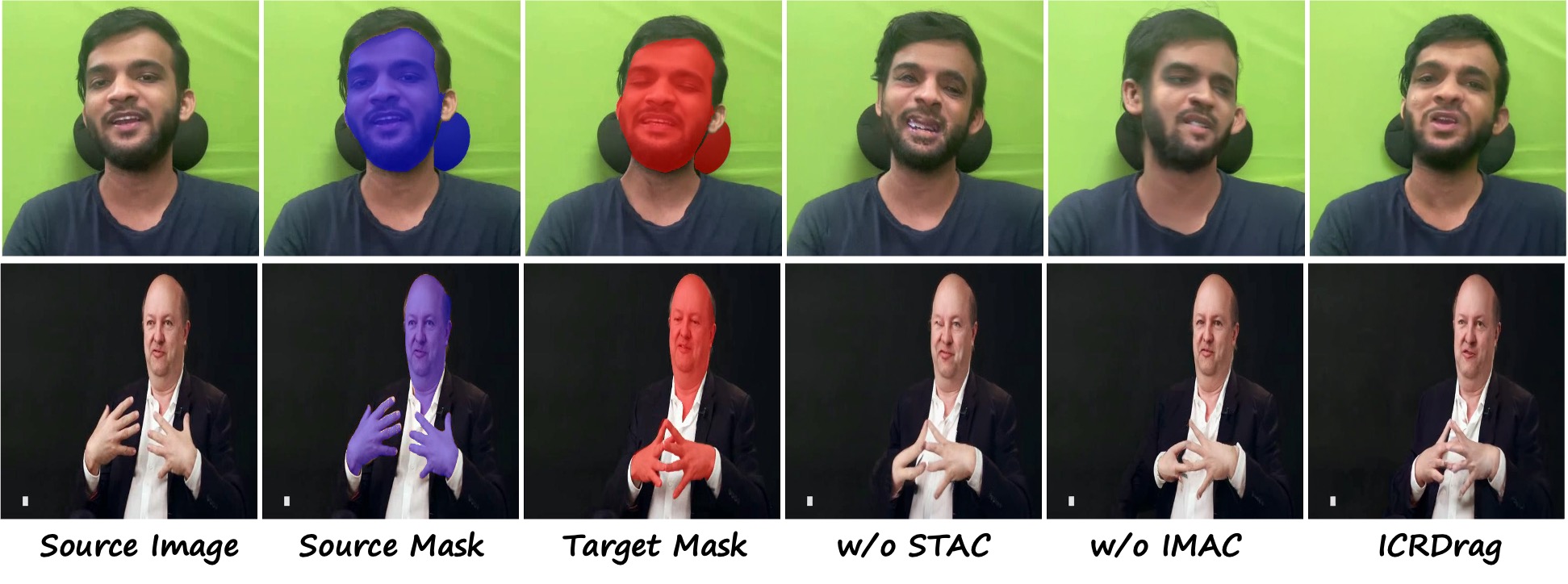}
  \caption{Visual results of ablation studies on our IMAC and STAC losses.}
  \label{fig:ablation}
\end{figure*}

\subsection{Ablation Study}
To evaluate Image-Mask Attention Consistency (IMAC) loss and Source-Target Attention Correspondence (STAC) loss, we conduct ablation studies by activating each loss separately during training. Figure \ref{fig:ablation} shows the visual results on our PRD benchmark. When IMAC is disabled, the model fails to strictly align the edit with the target mask, leading to noticeable structural distortions or misaligned boundaries. When STAC is disabled, the model struggles to preserve fine-grained details from the source image. Textures, patterns, or identity-specific features get altered or lost during the dragging process. The full model with both losses activated achieves the best performance, producing edits that are both precisely mask-aligned and visually faithful to the source. Additionally, the quantitative results of ablation study are presented in supplementary.

\subsection{Analysis of Attention Map}
\label{sec:attention_analysis}

\begin{figure}[t!]
    \centering
    \includegraphics[width=.99\textwidth]{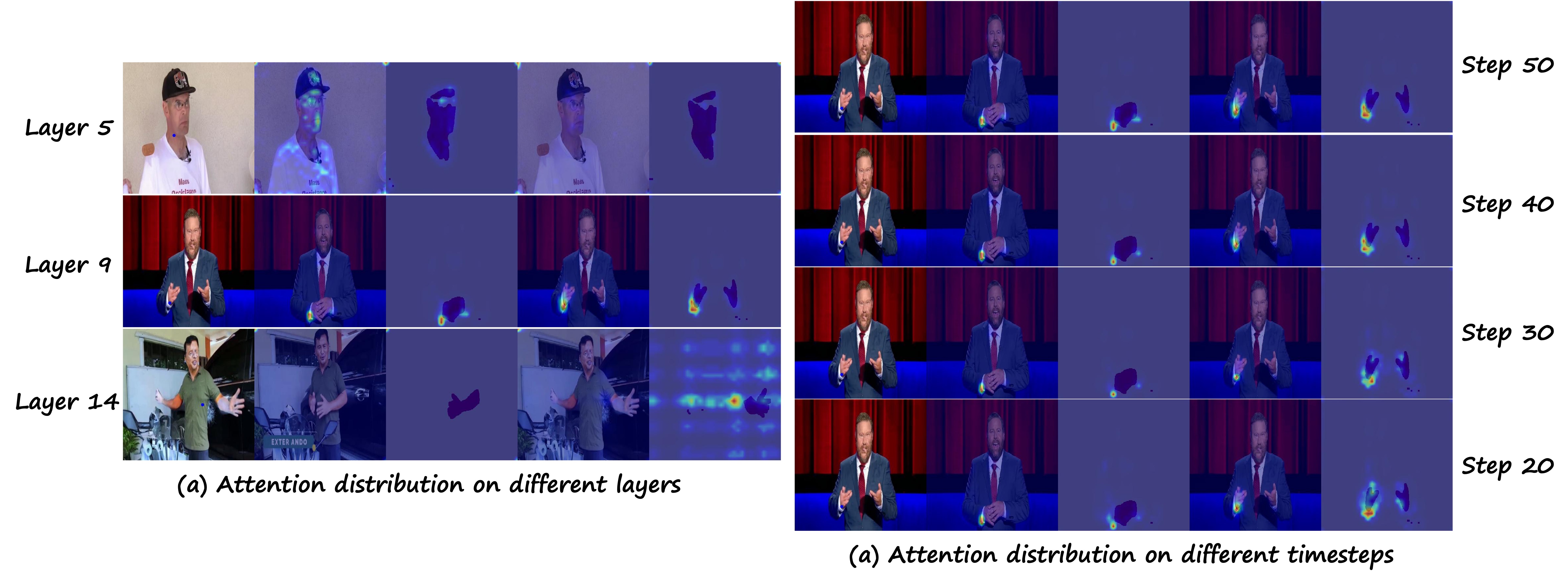}
    \caption{(a) Visualization of attention maps for a target patch across different transformer layers (NextDiT has 16 layers in total).
    (b) Attention maps from a middle transformer layer at different denoising timesteps.
    }
    \label{fig:attention_timesteps}
\end{figure}

We further analyze the attention maps of ICRDrag across transformer layers and denoising timesteps, as shown in Figure~\ref{fig:attention_timesteps}. The attention maps reveal a progressive transition from source-content localization in early layers, to source-target context integration in middle layers, and finally to spatial refinement in late layers. They also remain relatively stable through the whole denoising process, suggesting that source-target and image-mask correspondences are established early and consistently guide generation. These observations motivate applying IMAC and STAC losses to the middle transformer layers (7--11) and activating them through the whole denoising process.

\section{Conclusion}

In this paper, we have proposed In-context Region-based Drag (ICRDrag), which enhances the performance of region-based drag using in-context learning framework equipped with Image-Mask Attention Consistency (IMAC) and Source-Target Attention Correspondence (STAC) regularization. Additionally, we have constructed Paired Region Dataset (PRD) to further advance the research in the realm of drag-style editing. Experiments have shown that ICRDrag achieves superior performance in accuracy, detail, and realism, which demonstrates the potential of in-context drag-style editing model trained with large-scale data. 

\section*{Acknowledgements}
The work was supported by the National Natural Science Foundation of China (Grant No. 62471287).

% ---- Bibliography ----
%
% BibTeX users should specify bibliography style 'splncs04'.
% References will then be sorted and formatted in the correct style.
%
\bibliographystyle{splncs04}
\bibliography{main}

@String(CVPR  = {IEEE Conf. Comput. Vis. Pattern Recog.})

@String(ICCV  = {Int. Conf. Comput. Vis.})

@String(ECCV  = {Eur. Conf. Comput. Vis.})

@String(NeurIPS = {Adv. Neural Inform. Process. Syst.})

@String(ICML  = {Int. Conf. Mach. Learn.})

@String(ICLR  = {Int. Conf. Learn. Represent.})

@String(AAAI  = {AAAI})

@String(IJCAI = {IJCAI})

@String(ACMMM = {ACM Int. Conf. Multimedia})

@inproceedings{sohl2015deep,
  title={Deep unsupervised learning using nonequilibrium thermodynamics},
  author={Sohl-Dickstein, Jascha and Weiss, Eric and Maheswaranathan, Niru and Ganguli, Surya},
  booktitle={ICML},
  year={2015},
}

@inproceedings{ho2020denoising,
  title={Denoising diffusion probabilistic models},
  author={Ho, Jonathan and Jain, Ajay and Abbeel, Pieter},
  booktitle={NeurIPS},
  year={2020}
}

@inproceedings{pan2023drag,
  title={Drag your gan: Interactive point-based manipulation on the generative image manifold},
  author={Pan, Xingang and Tewari, Ayush and Leimk{\"u}hler, Thomas and Liu, Lingjie and Meka, Abhimitra and Theobalt, Christian},
  booktitle={ACM SIGGRAPH},
  year={2023}
}

@inproceedings{shi2023dragdiffusion,
  author={Shi, Yujun and Xue, Chuhui and Liew, Jun Hao and Pan, Jiachun and Yan, Hanshu and Zhang, Wenqing and F. Tan, Vincent Y. and Bai, Song},
  booktitle={CVPR}, 
  title={DragDiffusion: Harnessing Diffusion Models for Interactive Point-Based Image Editing}, 
  year={2024},
}

@inproceedings{mou2024dragon,
  title = {DragonDiffusion: Enabling Drag-style Manipulation on Diffusion Models},
  author = {Chong Mou and Xintao Wang and Jiechong Song and Ying Shan and Jian Zhang},
  booktitle = {ICLR},
  year = {2024}
}

@article{ling2023freedrag,
  title={Freedrag: Point tracking is not you need for interactive point-based image editing},
  author={Ling, Pengyang and Chen, Lin and Zhang, Pan and Chen, Huaian and Jin, Yi},
  journal={arXiv preprint arXiv:2307.04684},
  year={2023}
}

@article{Mou2024DiffEditorBA,
  title={DiffEditor: Boosting Accuracy and Flexibility on Diffusion-Based Image Editing},
  author={Chong Mou and Xintao Wang and Jie Song and Ying Shan and Jian Zhang},
  journal={CVPR},
  year={2024},
}

@inproceedings{jiang2024clipdrag,
 author = {Jiang, Ziqi and Wang, Zhen and Chen, Long},
 booktitle = {ICLR},
 title = {CLIPDrag: Combining Text-based and Drag-based Instructions for Image Editing},
 year = {2025}
}

@article{chen2024adaptivedrag,
  title={Adaptivedrag: Semantic-driven dragging on diffusion-based image editing},
  author={Chen, DuoSheng and Chen, Binghui and Geng, Yifeng and Bo, Liefeng},
  journal={arXiv preprint arXiv:2410.12696},
  year={2024}
}

@inproceedings{
koo2025flowdrag,
title={FlowDrag: 3D-aware Drag-based Image Editing with Mesh-guided Deformation Vector Flow Fields},
author={Gwanhyeong Koo and Sunjae Yoon and Younghwan Lee and Ji Woo Hong and Chang D. Yoo},
booktitle={ICML},
year={2025}
}

@inproceedings{
pu2026dragging,
title={Dragging with Geometry: From Pixels to Geometry-Guided Image Editing},
author={Xinyu Pu and Hongsong Wang and Jie Gui and Pan Zhou},
booktitle={ICLR},
year={2026}
}

@inproceedings{luo2023readout,
  author={Luo, Grace and Darrell, Trevor and Wang, Oliver and Goldman, Dan B and Holynski, Aleksander},
  booktitle={CVPR}, 
  title={Readout Guidance: Learning Control from Diffusion Features}, 
  year={2024},
}

@article{xu2023unifying,
  title={Unifying flow, stereo and depth estimation},
  author={Xu, Haofei and Zhang, Jing and Cai, Jianfei and Rezatofighi, Hamid and Yu, Fisher and Tao, Dacheng and Geiger, Andreas},
  journal={TPAMI},
  year={2023},
}

@inproceedings{zhang2018unreasonable,
  title={The unreasonable effectiveness of deep features as a perceptual metric},
  author={Zhang, Richard and Isola, Phillip and Efros, Alexei A and Shechtman, Eli and Wang, Oliver},
  booktitle={CVPR},
  year={2018}
}

@INPROCEEDINGS{nguyen2024edit,
  author={Nguyen, Thao and Ojha, Utkarsh and Li, Yuheng and Liu, Haotian and Lee, Yong Jae},
  booktitle={CVPR}, 
  title={Edit One for All: Interactive Batch Image Editing}, 
  year={2024},
}

@inproceedings{kim2022diffusionclip,
  title={Diffusionclip: Text-guided diffusion models for robust image manipulation},
  author={Kim, Gwanghyun and Kwon, Taesung and Ye, Jong Chul},
  booktitle={CVPR},
  year={2022}
}

@inproceedings{xu2023versatile,
  title={Versatile diffusion: Text, images and variations all in one diffusion model},
  author={Xu, Xingqian and Wang, Zhangyang and Zhang, Gong and Wang, Kai and Shi, Humphrey},
  booktitle={ICCV},
  year={2023}
}

@inproceedings{zhang2023adding,
  title={Adding conditional control to text-to-image diffusion models},
  author={Zhang, Lvmin and Rao, Anyi and Agrawala, Maneesh},
  booktitle={ICCV},
  year={2023}
}

@article{mou2023t2i,
  title={T2i-adapter: Learning adapters to dig out more controllable ability for text-to-image diffusion models},
  author={Mou, Chong and Wang, Xintao and Xie, Liangbin and Wu, Yanze and Zhang, Jian and Qi, Zhongang and Shan, Ying and Qie, Xiaohu},
  journal={arXiv preprint arXiv:2302.08453},
  year={2023}
}

@inproceedings{karnewar2023holodiffusion,
  title={Holodiffusion: Training a 3D diffusion model using 2D images},
  author={Karnewar, Animesh and Vedaldi, Andrea and Novotny, David and Mitra, Niloy J},
  booktitle={CVPR},
  year={2023}
}

@inproceedings{Loshchilov2017DecoupledWD,
  title={Decoupled Weight Decay Regularization},
  author={Ilya Loshchilov and Frank Hutter},
  booktitle={ICLR},
  year={2017},
}

@inproceedings{hou2024easydrag,
  title={EasyDrag: Efficient Point-based Manipulation on Diffusion Models},
  author={Hou, Xingzhong and Liu, Boxiao and Zhang, Yi and Liu, Jihao and Liu, Yu and You, Haihang},
  booktitle={CVPR},
  year={2024}
}

@inproceedings{liu2024drag,
  title={Drag your noise: Interactive point-based editing via diffusion semantic propagation},
  author={Liu, Haofeng and Xu, Chenshu and Yang, Yifei and Zeng, Lihua and He, Shengfeng},
  booktitle={CVPR},
  year={2024}
}

@inproceedings{zhang2024gooddrag,
    title={GoodDrag: Towards Good Practices for Drag Editing with Diffusion Models},
    author={Zewei Zhang and Huan Liu and Jun Chen and Xiangyu Xu},
    booktitle={ICLR},
    year={2025},
}

@inproceedings{cui2024stabledrag,
author="Cui, Yutao
and Zhao, Xiaotong
and Zhang, Guozhen
and Cao, Shengming
and Ma, Kai
and Wang, Limin",
title="StableDrag: Stable Dragging for Point-Based Image Editing",
booktitle={ECCV},
year={2025},
}

@inproceedings{li2024dragapart,
  title={DragAPart: Learning a Part-Level Motion Prior for Articulated Objects},
  author={Ruining Li and Chuanxia Zheng and Christian Rupprecht and Andrea Vedaldi},
  booktitle={ECCV},
  year={2024}
}

@inproceedings{
        zhao2024fastdrag,
        title={FastDrag: Manipulate Anything in One Step},
        author={Xuanjia Zhao and Jian Guan and Congyi Fan and Dongli Xu and Youtian Lin and Haiwei Pan and Pengming Feng},
        booktitle={NeurIPS},
        year={2024}
}

@inproceedings{
cui2024localize,
title={Localize, Understand, Collaborate: Semantic-Aware Dragging via Intention Reasoner},
author={Xing Cui and Pei Pei Li and Zekun Li and Xuannan Liu and Yueying Zou and Zhaofeng He},
booktitle={NeurIPS},
year={2024}
}

@inproceedings{avrahami2024diffuhaul,
  author = {Avrahami, Omri and Gal, Rinon and Chechik, Gal and Fried, Ohad and Lischinski, Dani and Vahdat, Arash and Nie, Weili},
  title = {DiffUHaul: A Training-Free Method for Object Dragging in Images},
  year = {2024},
  booktitle = {ACM SIGGRAPH Asia},
}

@inproceedings{
combing,
title={{CLIPD}rag: Combining Text-based and Drag-based Instructions for Image Editing},
author={Ziqi Jiang and Zhen Wang and Long Chen},
booktitle={ICLR},
year={2025}
}

@inproceedings{instantdrag,
      title     = {{InstantDrag: Improving Interactivity in Drag-based Image Editing}},
      author    = {Shin, Joonghyuk and Choi, Daehyeon and Park, Jaesik},
      booktitle = {ACM SIGGRAPH Asia},
      year      = {2024},
}

@inproceedings{ling2021editgan,
  title={Editgan: High-precision semantic image editing},
  author={Ling, Huan and Kreis, Karsten and Li, Daiqing and Kim, Seung Wook and Torralba, Antonio and Fidler, Sanja},
  booktitle={NeurIPS},
  year={2021}
}

@inproceedings{regiondrag,
  author    = {Jingyi Lu and Xinghui Li and Kai Han},
  title     = {RegionDrag: Fast Region-Based Image Editing with Diffusion Models},
  booktitle = {ECCV},
  year      = {2024},
}

@inproceedings{
shi2024instadrag,
title={LightningDrag: Lightning Fast and Accurate Drag-based Image Editing Emerging from Videos},
author={Yujun Shi and Jun Hao Liew and Hanshu Yan and Vincent Y. F. Tan and Jiashi Feng},
booktitle={ICML},
year={2025}
}

@inproceedings{wang2023context,
  title={In-context learning unlocked for diffusion models},
  author={Wang, Zhendong and Jiang, Yifan and Lu, Yadong and He, Pengcheng and Chen, Weizhu and Wang, Zhangyang and Zhou, Mingyuan and others},
  booktitle={NeurIPS},
  year={2023}
}

@inproceedings{dong2022survey,
    title = "A Survey on In-context Learning",
    author = "Dong, Qingxiu  and
      Li, Lei  and
      Dai, Damai  and
      Zheng, Ce  and
      Ma, Jingyuan  and
      Li, Rui  and
      Xia, Heming  and
      Xu, Jingjing  and
      Wu, Zhiyong  and
      Chang, Baobao  and
      Sun, Xu  and
      Li, Lei  and
      Sui, Zhifang",
    booktitle = {EMNLP},
    year = "2024",
}

@article{huang2024context,
  title={In-context lora for diffusion transformers},
  author={Huang, Lianghua and Wang, Wei and Wu, Zhi-Fan and Shi, Yupeng and Dou, Huanzhang and Liang, Chen and Feng, Yutong and Liu, Yu and Zhou, Jingren},
  journal={arXiv preprint arXiv:2410.23775},
  year={2024}
}

@inproceedings{selfsupericl,
    title = "Improving In-Context Few-Shot Learning via Self-Supervised Training",
    author = "Chen, Mingda  and
      Du, Jingfei  and
      Pasunuru, Ramakanth  and
      Mihaylov, Todor  and
      Iyer, Srini  and
      Stoyanov, Veselin  and
      Kozareva, Zornitsa",
    booktitle = {ACL},
    year = {2022},
}

@inproceedings{picl,
  author       = {Yuxian Gu and
                  Li Dong and
                  Furu Wei and
                  Minlie Huang},
  title        = {Pre-Training to Learn in Context},
  booktitle    = {ACL},
  year         = {2023}
}

@inproceedings{
Shi2023iclm,
title={In-Context Pretraining: Language Modeling Beyond Document Boundaries},
author={Weijia Shi and Sewon Min and Maria Lomeli and Chunting Zhou and Margaret Li and Xi Victoria Lin and Noah A. Smith and Luke Zettlemoyer and Wen-tau Yih and Mike Lewis},
booktitle={ICLR},
year={2024},
}

@inproceedings{
zhuo2024lumina,
title={Lumina-Next : Making Lumina-T2X Stronger and Faster with Next-DiT},
author={Le Zhuo and Ruoyi Du and Han Xiao and Yangguang Li and Dongyang Liu and Rongjie Huang and Wenze Liu and Xiangyang Zhu and Fu-Yun Wang and Zhanyu Ma and Xu Luo and Zehan Wang and Kaipeng Zhang and Lirui Zhao and Si Liu and Xiangyu Yue and Wanli Ouyang and Yu Qiao and Hongsheng Li and Peng Gao},
booktitle={NeurIPS},
year={2024}
}

@inproceedings{nan2024openvid,
 author = {Nan, Kepan and Xie, Rui and Zhou, Penghao and Fan, Tiehan and Yang, Zhenheng and Chen, Zhijie and Li, Xiang and Yang, Jian and Tai, Ying},
 booktitle = {ICLR},
 title = {OpenVid-1M: A Large-Scale High-Quality Dataset for Text-to-video Generation},
 year = {2025}
}

@article{matsunaga2022fine,
  title={Fine-grained Image Editing by Pixel-wise Guidance Using Diffusion Models},
  author={Matsunaga, Naoki and Ishii, Masato and Hayakawa, Akio and Suzuki, Kenji and Narihira, Takuya},
  journal={AI for Content Creation workshop at CVPR},
  year={2022}
}

@inproceedings{kirillov2023segment,
  title={Segment anything},
  author={Kirillov, Alexander and Mintun, Eric and Ravi, Nikhila and Mao, Hanzi and Rolland, Chloe and Gustafson, Laura and Xiao, Tete and Whitehead, Spencer and Berg, Alexander C and Lo, Wan-Yen and others},
  booktitle={ICCV},
  year={2023}
}

@inproceedings{esser2024scaling,
  title={Scaling rectified flow transformers for high-resolution image synthesis},
  author={Esser, Patrick and Kulal, Sumith and Blattmann, Andreas and Entezari, Rahim and M{\"u}ller, Jonas and Saini, Harry and Levi, Yam and Lorenz, Dominik and Sauer, Axel and Boesel, Frederic and others},
  booktitle={ICML},
  year={2024}
}

@inproceedings{
lipman2022flow,
title={Flow Matching for Generative Modeling},
author={Yaron Lipman and Ricky T. Q. Chen and Heli Ben-Hamu and Maximilian Nickel and Matthew Le},
booktitle={ICLR},
year={2023}
}

@inproceedings{hu2022lora,
  title={Lora: Low-rank adaptation of large language models.},
  author={Hu, Edward J and Shen, Yelong and Wallis, Phillip and Allen-Zhu, Zeyuan and Li, Yuanzhi and Wang, Shean and Wang, Lu and Chen, Weizhu and others},
  booktitle={ICLR},
  year={2022}
}

@inproceedings{bengio2009curriculum,
  title={Curriculum learning},
  author={Bengio, Yoshua and Louradour, J{\'e}r{\^o}me and Collobert, Ronan and Weston, Jason},
  booktitle={ICML},
  year={2009}
}

@article{wang2021survey,
  title={A survey on curriculum learning},
  author={Wang, Xin and Chen, Yudong and Zhu, Wenwu},
  journal={TPAMI},
  volume={44},
  number={9},
  pages={4555--4576},
  year={2021},
}

@inproceedings{pentina2015curriculum,
  title={Curriculum learning of multiple tasks},
  author={Pentina, Anastasia and Sharmanska, Viktoriia and Lampert, Christoph H},
  booktitle={CVPR},
  year={2015}
}

@inproceedings{hacohen2019power,
  title={On the power of curriculum learning in training deep networks},
  author={Hacohen, Guy and Weinshall, Daphna},
  booktitle={ICML},
  year={2019},
}

@article{li2023semantic,
  title={Semantic-SAM: Segment and Recognize Anything at Any Granularity},
  author={Li, Feng and Zhang, Hao and Sun, Peize and Zou, Xueyan and Liu, Shilong and Yang, Jianwei and Li, Chunyuan and Zhang, Lei and Gao, Jianfeng},
  journal={arXiv preprint arXiv:2307.04767},
  year={2023}
}

@article{ravi2024sam2,
  title={SAM 2: Segment Anything in Images and Videos},
  author={Ravi, Nikhila and Gabeur, Valentin and Hu, Yuan-Ting and Hu, Ronghang and Ryali, Chaitanya and Ma, Tengyu and Khedr, Haitham and R{\"a}dle, Roman and Rolland, Chloe and Gustafson, Laura and Mintun, Eric and Pan, Junting and Alwala, Kalyan Vasudev and Carion, Nicolas and Wu, Chao-Yuan and Girshick, Ross and Doll{\'a}r, Piotr and Feichtenhofer, Christoph},
  journal={arXiv preprint arXiv:2408.00714},
  year={2024}
}

@inproceedings{wang2025training,
    title={Training-free dense-aligned diffusion guidance for modular conditional image synthesis},
    author={Wang, Zixuan and Peng, Duo and Chen, Feng and Yang, Yuwei and Lei, Yinjie},
    booktitle={CVPR},
    year={2025}
}

@inproceedings{yan2025eedit,
  author={Yan, Zexuan and Ma, Yue and Zou, Chang and Chen, Wenteng and Chen, Qifeng and Zhang, Linfeng},
  booktitle={ICCV}, 
  title={EEdit: Rethinking the Spatial and Temporal Redundancy for Efficient Image Editing}, 
  year={2025},
}

@inproceedings{zhang2025framepainter,
  author={Zhang, Yabo and Zhou, Xinpeng and Zeng, Yihan and Xu, Hang and Li, Hui and Zuo, Wangmeng},
  booktitle={ICCV}, 
  title={FramePainter: Endowing Interactive Image Editing with Video Diffusion Priors}, 
  year={2025},
}

@article{xia2024dreamomni,
  title={DreamOmni: Unified Image Generation and Editing},
  author={Xia, Bin and Zhang, Yuechen and Li, Jingyao and Wang, Chengyao and Wang, Yitong and Wu, Xinglong and Yu, Bei and Jia, Jiaya},
  journal={arXiv preprint arXiv:2412.17098},
  year={2024}
}

@inproceedings{cai2024auto,
  title={Auto DragGAN: Editing the Generative Image Manifold in an Autoregressive Manner},
  author={Cai, Pengxiang and Liu, Zhiwei and Zhu, Guibo and Niu, Yunfang and Wang, Jinqiao},
  booktitle={ACMMM},
  year={2024}
}

@inproceedings{choi2025dragtext,
  title={Dragtext: Rethinking Text Embedding in Point-Based Image Editing},
  author={Choi, Gayoon and Jeong, Taejin and Hong, Sujung and Hwang, Seong Jae},
  booktitle={WACV},
  year={2025},
}

@inproceedings{xia2025draglora,
  title={DragLoRA: Online Optimization of LoRA Adapters for Drag-based Image Editing in Diffusion Model},
  author={Xia, Siwei and Sun, Li and Sun, Tiantian and Li, Qingli},
  booktitle={ICML},
  year={2025}
}

@inproceedings{zhou2025dragnext,
  title={DragNeXt: Rethinking Drag-Based Image Editing},
  author={Yuan Zhou and Junbao Zhou and Qingshan Xu and Kesen Zhao and Yuxuan Wang and Hao Fei and Richang Hong and Hanwang Zhang},
  booktitle={AAAI},
  year={2025}
}

@inproceedings{
nie2023blessing,
title={The Blessing of Randomness: {SDE} Beats {ODE} in General Diffusion-based Image Editing},
author={Shen Nie and Hanzhong Allan Guo and Cheng Lu and Yuhao Zhou and Chenyu Zheng and Chongxuan Li},
booktitle={ICLR},
year={2024}
}

@inproceedings{
yin2025lazydrag,
title={LazyDrag: Enabling Stable Drag-Based Editing on Multi-Modal Diffusion Transformers via Explicit Correspondence},
author={Zixin Yin and Xili Dai and Duomin Wang and Xianfang Zeng and Lionel Ni and Gang YU and Heung-Yeung Shum},
booktitle={ICLR},
year={2026}
}

@inproceedings{liao2025directdrag,
    author    = {Liao, Sheng-Hao and Chen, Shang-Fu and Huang, Tai-Ming and Cheng, Wen-Huang and Hua, Kai-Lung},
    title     = {DirectDrag: High-Fidelity, Mask-Free, Prompt-Free Drag-based Image Editing via Readout-Guided Feature Alignment},
    booktitle = {WACV},
    year      = {2026},
}

@article{he2025contextdrag,
  title={ContextDrag: Precise Drag-Based Image Editing via Context-Preserving Token Injection and Position-Consistent Attention},
  author={He, Huiguo and Yan, Pengyu and Yi, Ziqi and Zhong, Weizhi and Liu, Zheng and Tang, Yejun and Yang, Huan and Gai, Kun and Li, Guanbin and Jin, Lianwen},
  journal={arXiv preprint arXiv:2512.08477},
  year={2025}
}

@inproceedings{yang2025attentiondrag,
  title={AttentionDrag: Exploiting Latent Correlation Knowledge in Pre-trained Diffusion Models for Image Editing},
  author={Biao Yang and Muqi Huang and Yuhui Zhang and Yun Xiong and Kun Zhou and Xi Chen and Shiyang Zhou and Huishuai Bao and Chuan Li and Feng Shi and Hualei Liu},
  booktitle={IJCAI},
  year={2025}
}

@inproceedings{
zhou2025dragflow,
title={DragFlow: Unleashing DiT Priors with Region-Based Supervision for Drag Editing},
author={Zihan Zhou and Shilin Lu and Shuli Leng and Shaocong Zhang and Zhuming Lian and Xinlei Yu and Adams Wai-Kin Kong},
booktitle={ICLR},
year={2026}
}

@inproceedings{
baranchuk2021labelefficient,
title={Label-Efficient Semantic Segmentation with Diffusion Models},
author={Dmitry Baranchuk and Andrey Voynov and Ivan Rubachev and Valentin Khrulkov and Artem Babenko},
booktitle={ICLR},
year={2022}
}

@inproceedings{lu2025inpaint4drag,
 author    = {Jingyi Lu and Kai Han},
 title     = {Inpaint4Drag: Repurposing Inpainting Models for Drag-Based Image Editing via Bidirectional Warping},
 booktitle = {ICCV},
 year      = {2025},
}
\end{document}

% ---------------------------------------------------------------
% TODO REVIEW: Replace with your title
\title{Supplementary for In-context Region-based Drag: Drag Any Region to Any Shape} 

% TODO REVIEW: If the paper title is too long for the running head, you can set
% an abbreviated paper title here. If not, comment out.
% \titlerunning{Abbreviated paper title}

% TODO FINAL: Replace with your author list. 
% Include the authors' OCRID for the camera-ready version, if at all possible.
% Equal contribution mark
\newcommand{\equalcontrib}{\textsuperscript{\textdagger}}
\newcommand{\corresponding}{\textsuperscript{*}}

\author{Jiacheng Sui\equalcontrib \and
Tianyu Hao\orcidlink{0009-0000-6488-5985}\equalcontrib \and
Bingjie Gao\orcidlink{0000-0003-3622-7509} \and
Li Niu\orcidlink{0000-0003-1970-8634}\corresponding \and
Guangtao Zhai\orcidlink{0000-0001-8165-9322}}

% TODO FINAL: Replace with an abbreviated list of authors.
\authorrunning{J. Sui et al.}
% First names are abbreviated in the running head.
% If there are more than two authors, 'et al.' is used.

% TODO FINAL: Replace with your institution list.
\institute{
Shanghai Jiao Tong University\\
\email{ jcsui01@sjtu.edu.cn, htianyu429@gmail.com, whynothaha@sjtu.edu.cn, ustcnewly@sjtu.edu.cn, zhaiguangtao@sjtu.edu.cn }
}

\maketitle

In this supplementary material, we provide additional analyses and results organized as follows: In Section \ref{sec:attention_analysis}, we visualize and compare attention maps from models trained with and without IMAC or STAC.
In Section \ref{sec:ablation_quantitative}, we provide the quantitative metrics for the ablation study on IMAC and STAC. In Section \ref{sec:ablation_twostage}, we conduct ablation study on two-stage training strategy.
Section \ref{sec:implementation} presents implementation details. In Section \ref{sec:inference_time}, we analyze the computational efficiency of our ICRDrag compared to representative methods.
Section \ref{sec:prd} showcases additional examples from the Paired Region Dataset. Section \ref{sec:qualitative} presents more visual editing results produced by ICRDrag. In Section \ref{sec:ablation}, we provide further visual ablation results for ICRDrag. In Section \ref{sec:failurecases}, we present and analyze several failure cases of the editing process. Finally, Section \ref{sec:impacts} discusses the broader impacts potentially introduced by ICRDrag.

\section{Attention Map Analysis}
\label{sec:attention_analysis}

To gain deeper insights into how IMAC and STAC losses influence the model's internal representations, we visualize and compare attention maps from models trained with and without each loss. Figure~\ref{fig:attention_analysis} presents the results.

\subsection{IMAC Attention Map Analysis}
\label{sec:attention_imac}

\begin{figure}[t!]
    \centering
    \includegraphics[width=0.8\textwidth]{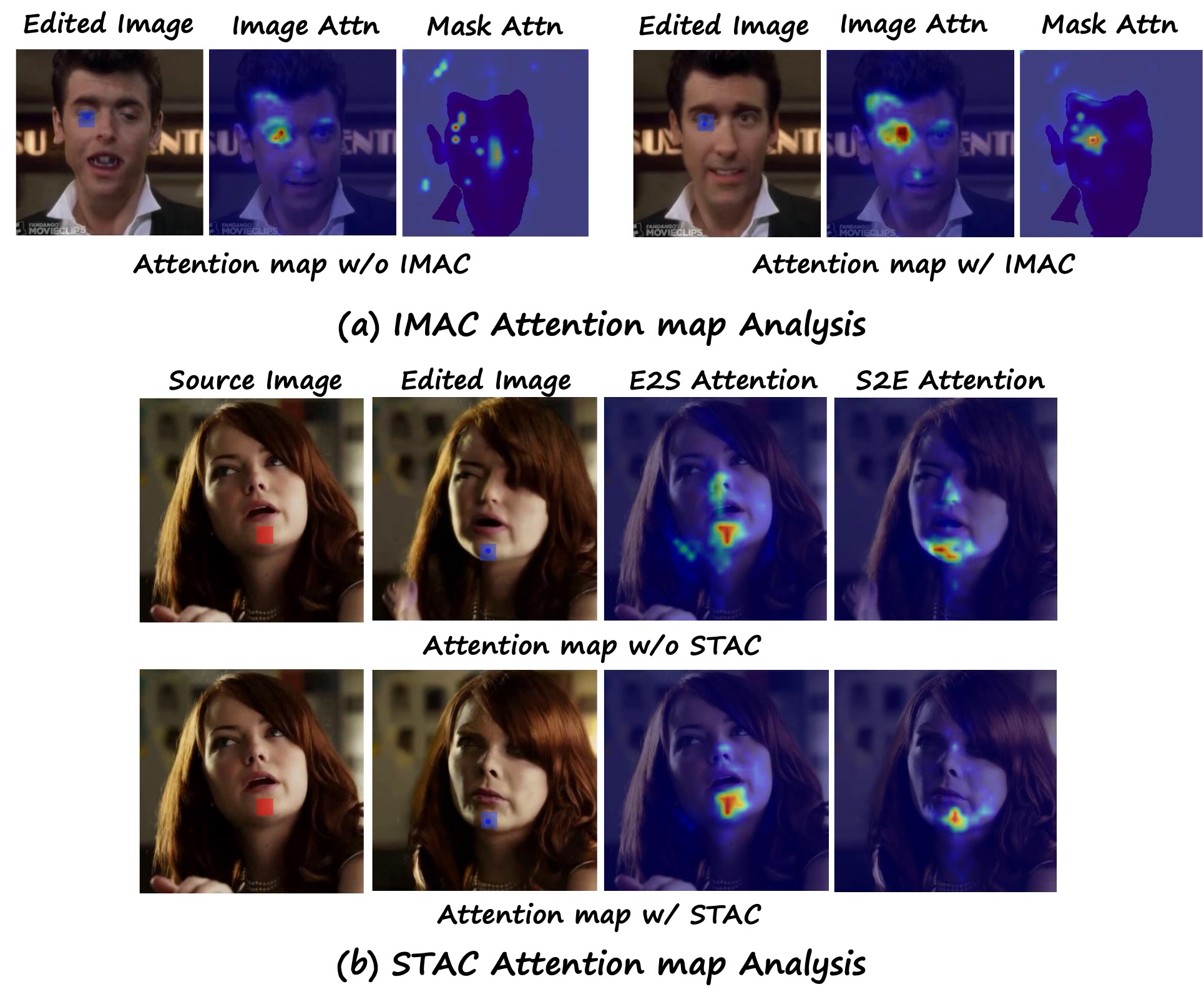}
    \caption{Attention map analysis. (a) IMAC attention maps: for a target patch (marked blue in "Edited Image" column), we visualize attention over source image ("Image Attn" column) and corresponding target mask patch's attention over source mask ("Mask Attn" column). Left: w/o IMAC, right: w/ IMAC. (b) STAC attention maps: for a target patch (marked blue in "Edited Image" column) and its corresponding source patch (marked red in "Source Image" column), we visualize attention from edited to source attention("E2S Attention" column) and from source to edited attention ("S2E Attention" column). Top: w/o STAC, bottom: w/ STAC.}
    \label{fig:attention_analysis}
\end{figure}

To understand how Image-Mask Attention Consistency (IMAC) influences cross-modal alignment, we visualize attention maps from models trained with and without this regularization. Figure~\ref{fig:attention_analysis}(a) presents a comparison result.

\paragraph{Visualization Setup.}
For a given query patch located in the target region of the edited image (marked in blue in the "Edited Image" column), we visualize two attention distributions. The "Image Attn" columns show the attention of this target patch over the source image. The "Mask Attn" columns show the attention of the corresponding target mask patch (at the same spatial location) over the source mask. The left group presents results from the ablation model without IMAC, while the right group shows our full ICRDrag model with IMAC activated.

\paragraph{Observation.}
In both models, we observe that the image attention remains relatively precise, focusing on the relevant region. This indicates that the model inherently learns to locate visual content reasonably well even without explicit regularization. However, a critical difference emerges in the mask attention. When IMAC is not activated, the mask attention exhibits poor alignment with the image attention — it disperses or focuses on incorrect regions, failing to mirror the spatial grounding provided by the image attention. This cross-modal discrepancy means that while the model knows where to look visually, it lacks the structural guidance from the mask to confirm that the attended region is semantically correct.

With IMAC activated, the mask attention becomes tightly aligned with the image attention. This alignment confirms that our IMAC regularization successfully enforces cross-modal consistency, ensuring that the visual features are drawn from the region that the mask identifies as semantically relevant. The resulting attention alignment explains the improved structural accuracy achieved by our full model.

\subsection{STAC Attention Map Analysis}
\label{sec:attention_stac}

To understand how Source-Target Attention Correspondence (STAC) influences bidirectional attention, we visualize attention maps from models trained with and without this regularization. Figure~\ref{fig:attention_analysis}(b) presents the results for a human pose editing task.

\paragraph{Visualization Setup.}
For a given pair of corresponding patches, we visualize two attention distributions. The "E2S Attention" column shows the attention from the target patch to the source image, \ie, where the target patch looks to gather visual content. The "S2E Attention" column shows the attention from the source patch to the edited image, \ie, where the source patch looks in return. The first row presents results from the ablation model without STAC, while the second row shows our full ICRDrag model with STAC activated.

\paragraph{Observation.}
When STAC is not activated, both attention maps exhibit undesirable spreading. The target patch's attention to the source image disperses across multiple regions beyond the intended corresponding source patch, including unrelated face parts. Similarly, the source patch's attention back to the target image is diffused and fails to focus on the correct target location. This bilateral diffusion indicates that without explicit regularization, the model lacks the mechanism to maintain focused correspondences, leading to the detail degradation and texture blurring observed in our visual ablation study.

With STAC activated, both directions exhibit strong focused attention. The target patch attends to the corresponding source patch, while the source patch attends back precisely to the corresponding target patch. This symmetric mutual attention creates a closed-loop correspondence that reinforces detail preservation throughout the generation process. The source patch no longer gets distracted by irrelevant regions, ensuring that fine-grained features are faithfully transferred.

These attention visualizations provide clear evidence of how our proposed losses shape the model's internal representations. IMAC aligns cross-modal attention to ensure structural grounding, while STAC establishes bidirectional correspondence to preserve fine-grained details. 

\begin{table*}[t]
  \centering
  \caption{Quantitative ablation study on IMAC and STAC.}
  \label{tab:ablation_quantitative}
  \begin{tabular}{c|ccccc}
    \toprule
       Method         & MSE $\downarrow$    & LPIPS$\downarrow$ & SSIM $\uparrow$& MD(RegionDrag)$\downarrow$ & MD(DragLoRA)$\downarrow$\\
    \midrule
    w/o Both & 0.0981 & 0.1979 & 0.5953 & 5.52 & 35.05 \\
    w/o STAC & 0.0825 & 0.1722 & 0.6121 & 3.92 & 24.06 \\
    w/o IMAC & 0.0934 & 0.1925 & 0.6073 & 4.77 & 26.72 \\
    ICRDrag   & \textbf{0.0735} & \textbf{0.1610} & \textbf{0.6284} & \textbf{3.66} & \textbf{22.34} \\
    \bottomrule
  \end{tabular}
\end{table*}

\section{Quantitative Ablation Study on IMAC and STAC}
\label{sec:ablation_quantitative}

We conduct quantitative ablation studies to evaluate the individual and combined contributions of our proposed losses: Image-Mask Attention Consistency (IMAC) and Source-Target Attention Correspondence (STAC). All models are trained on the PRD training set and evaluated on our PRD benchmark using four metrics: MSE, LPIPS~\cite{zhang2018unreasonable}, SSIM and Mean Distance(MD).

Table~\ref{tab:ablation_quantitative} presents the quantitative results. The baseline model (without IMAC and STAC) establishes a reference performance across all metrics. Adding IMAC alone improves all metrics, demonstrating that cross-modal attention alignment enhances editing accuracy and structural fidelity. Adding STAC alone yields comparable improvements, confirming that bidirectional attention correspondence contributes to better detail preservation.
When both losses are activated together, our full model achieves the best performance across all metrics, outperforming either individual loss. The improvements over each loss indicate that IMAC and STAC address complementary aspects of the dragging task. IMAC ensures structural alignment with masks, while STAC preserves fine-grained details through mutual correspondence.

\begin{table}[t!]
\centering
\caption{Ablation on two-stage training strategy. The two-stage curriculum strategy achieves the best performance by progressively increasing task difficulty.}
\label{tab:ablation_twostage}
\begin{tabular}{lccc}
\toprule
Training Strategy & MSE$\downarrow$ & LPIPS$\downarrow$ & SSIM$\uparrow$ \\
\midrule
Stage-1 only & 0.0807 & 0.1615 & 0.6233 \\
Stage-2 only & 0.0779 & 0.1628 & 0.6225 \\
Two-stage &  \textbf{0.0735} & \textbf{0.1610} & \textbf{0.6284}  \\
\bottomrule
\end{tabular}
\end{table}

\begin{figure}[t!]
    \centering
    \includegraphics[width=0.9\textwidth]{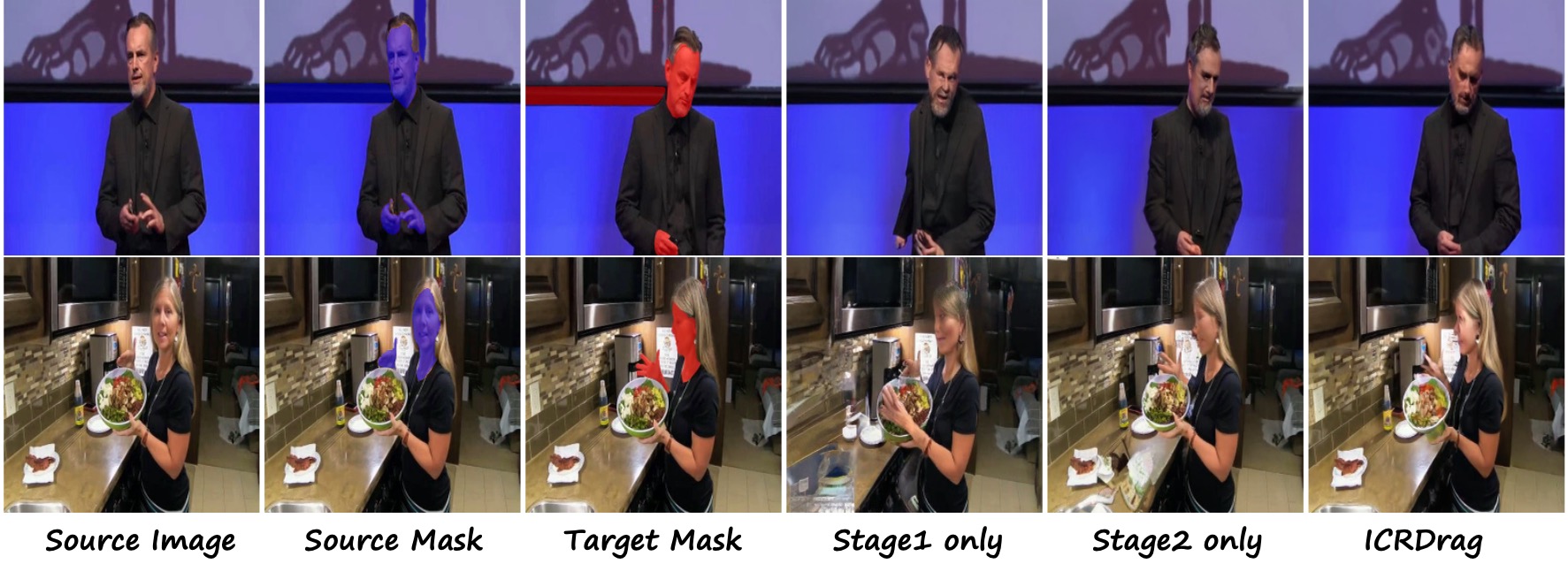}
    \caption{Visual comparison of training strategies. Stage1-only (complete masks) fails to coordinate natural movement and introduces artifacts. Stage2-only (incomplete masks) exhibits poor non-edited region preservation, including color shift (first) and object hallucination (second). Our two-stage strategy resolves both issues.}
    \label{fig:twostage_ablation}
\end{figure}

\section{Ablation Study on Two-stage Training Strategy}
\label{sec:ablation_twostage}

We conduct ablation studies to validate the effectiveness of our proposed two-stage curriculum training strategy. We compare three variants: (1) training from scratch on complete masks only (stage1-only), (2) training from scratch on incomplete masks only (stage2-only), and (3) our full two-stage strategy (pre-training on complete masks followed by fine-tuning on incomplete masks). All variants are evaluated on the PRD benchmark. As shown in Figure~\ref{fig:twostage_ablation}, both single-stage variants exhibit notable limitations.

\paragraph{Stage1-only Limitations.}
Training exclusively on complete masks leads to a critical distribution shift during inference. Since this variant is trained only on dense fully-separated masks, it learns to treat any gray region in the mask as areas that should remain unchanged. However, real-world user inputs are often incomplete. Only a few regions are separated out while the rest are filled with gray. This causes the model to misinterpret gray areas as strictly static, leading to editing conflicts. In the first example of Figure~\ref{fig:twostage_ablation}, the stage1-only model fails to make the body follow the head's rotation naturally, resulting in an unnatural and rigid edit. In the second example, this misinterpretation leads to visible artifacts in non-edited areas, such as distorted background structures.

\paragraph{Stage2-only Limitations.}
Training solely on incomplete masks exhibits a different problem: poor preservation of non-edited regions. Without the foundation established by pre-training on complete mask, the model struggles to understand which areas should remain unchanged. This leads to unintended modifications outside the editing region. In the first example, the stage2-only model introduces subtle color shift in the background. More severely, in the complex scene of the second example, the model hallucinates objects that do not exist in the source image, such as an additional item appearing on the table. We attribute this to over-reasoning induced by training only on sparse masks. The model learns to infer too aggressively about unmarked regions, attempting to "complete" them in ways that deviate from the ground-truth.

Our full two-stage strategy addresses both limitations. Pretraining on complete masks establishes a strong prior for understanding mask-image relationships and preserving non-edited regions. Fine-tuning on incomplete masks then adapts the model to real-world sparse inputs while retaining this prior. The resulting edits are both structurally accurate and faithful to non-edited areas, as shown in the ICRDrag column.

We also perform ablation study on two-stage training strategy and report the results in Table \ref{tab:ablation_twostage}. It can be seen that ICRDrag with two-stage training strategy achieves the best performance by progressively increasing task difficulty. These results demonstrate that our two-stage curriculum training strategy is essential for balancing edit accuracy with non-edited region preservation, enabling robust generalization to real-world incomplete mask inputs.

\section{Implementation Details}
\label{sec:implementation}

\paragraph{Model} Our ICRDrag takes Next-DiT \cite{zhuo2024lumina} as our backbone. Nevertheless, our proposed ICRDrag is compatible with any model based on the DiT architecture.

\paragraph{Training} During the training stage 1, we train the model for 60,000 steps, with batch size 2 and learning rate $1 \times 10^{-4}$. While for training stage 2, we train the model for another 2,000 steps, with batch size 1 and learning rate $5 \times 10^{-5}$. During both stages, gradient checkpointing is applied to save VRAM and AdamW \cite{Loshchilov2017DecoupledWD} optimizer with $\beta_1 = 0.9, \beta_2 = 0.999$ is employed. All the experiments running ICRDrag in this work are done on 4 NVIDIA Pro6000 GPU with 96GB VRAM.

\paragraph{Inference} When editing images during inference period, we apply FlowMatching \cite{lipman2022flow} scheduler with 50 steps. 

\section{Inference Time Analysis}
\label{sec:inference_time}

As illustrated in Table \ref{tab:inference_time}, we analyze the computational efficiency of our ICRDrag framework compared to representative methods. All experiments are conducted on a single NVIDIA A6000 GPU with input resolution 512×512, averaging over 100 test images.

\begin{table}[t!]
\centering
\caption{Average inference time per image on NVIDIA A6000. Our method achieves competitive efficiency while maintaining high editing quality.}
\label{tab:inference_time}
\small
\begin{tabular}{lcccccc}
\toprule
Metric & DragDiffusion & DragLoRA & GoodDrag & Inpaint4Drag & RegionDrag & ICRDrag \\
\midrule
Time(s) $\downarrow$ & 157.50 & 87.94 & 87.78 & 0.15 & 0.85 & 35.62 \\
\bottomrule
\end{tabular}

\end{table}

The inference time of our method reflects a deliberate design choice: we prioritize editing quality and spatial precision over raw speed. The 35.62s inference time enables our model to perform fine-grained attention-based reasoning across multiple modalities, which is essential for accurate region-based dragging. In contrast, methods like RegionDrag achieve faster inference through simpler copy-paste operations but struggle with boundary consistency and complex deformations, as shown in our qualitative results.

For practical applications, this trade-off is often acceptable — users typically value editing accuracy over a few seconds of waiting time. Moreover, our inference pipeline is amenable to optimization techniques such as step distillation or parallel sampling, which could further reduce runtime without compromising quality. We leave such acceleration strategies for future work.

\begin{figure*}[t!]
  \centering
  \includegraphics[width=\textwidth]{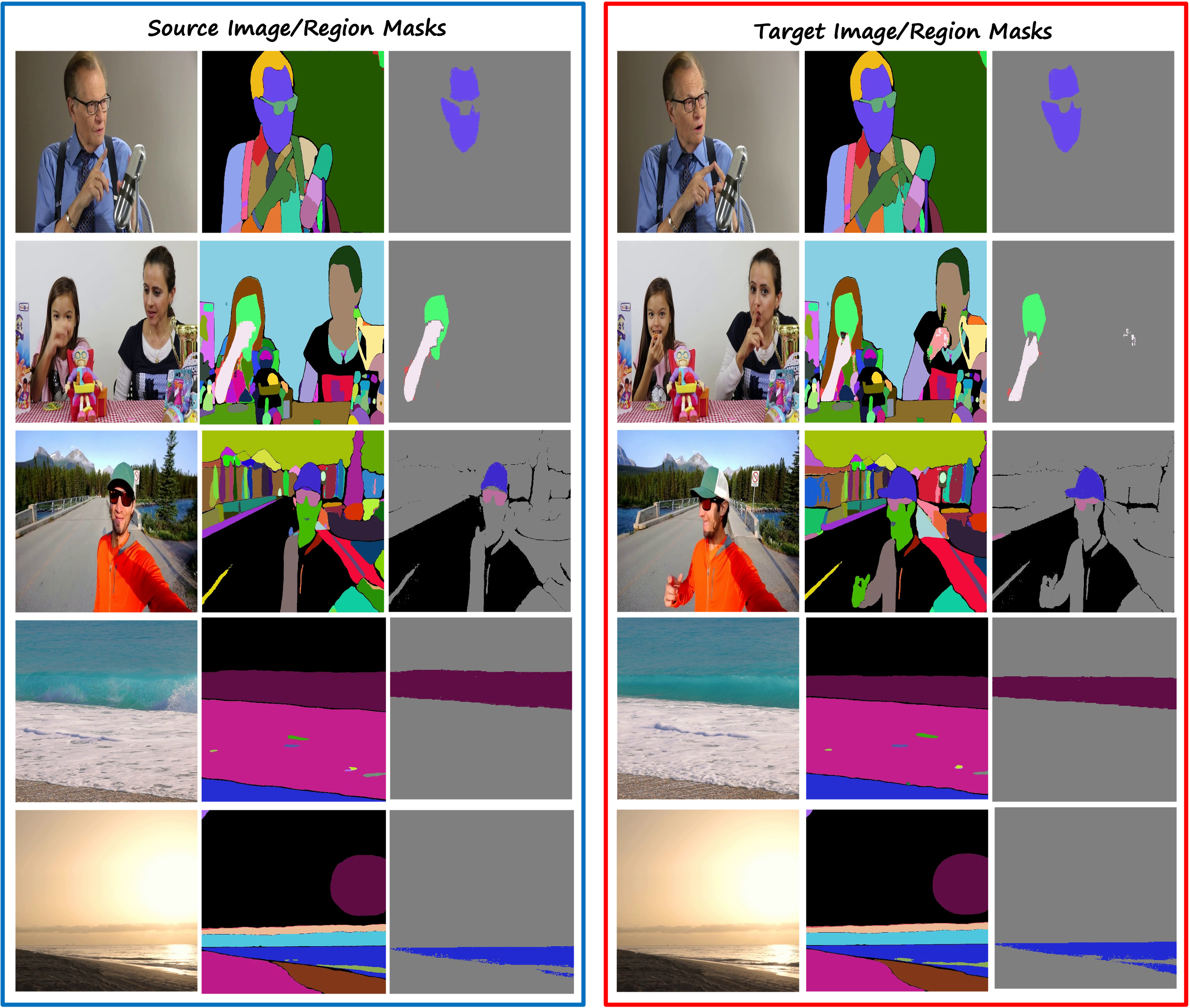}
  \caption{More examples of Paired Region Dataset. The left side of the image displays the source image along with its complete and incomplete region masks, while the right side shows the corresponding target image and its associated complete and incomplete region masks.
  }
  \label{fig:dataset_examples}
\end{figure*}

\section{Paired Region Dataset (PRD) Examples}
\label{sec:prd}
As shown in Figure \ref{fig:dataset_examples}, we present additional instances from the Paired Region Dataset. For more details on the construction of PRD, please refer to Section 5 in our paper. The left side of the image displays the source image along with its complete and incomplete region masks, while the right side shows the corresponding target image and its associated complete and incomplete region masks.

\section{More Qualitative Results}
\label{sec:qualitative}

\begin{figure*} [t]
  \centering
  \includegraphics[width=\textwidth]{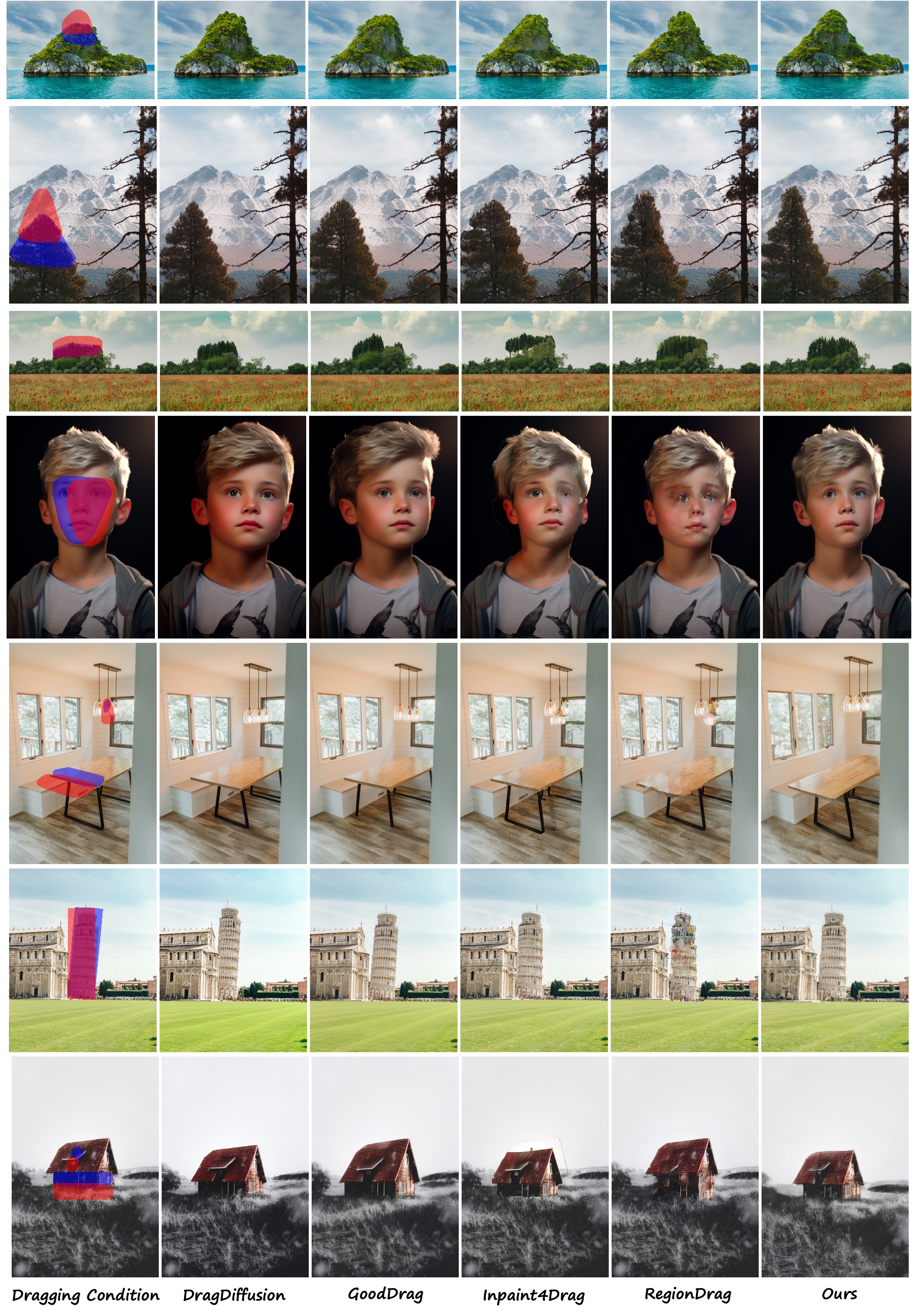}
  \caption{More visual qualitative results on DragBench-DR. In both the left and right subfigures, the visualizations from left to right are as follows: the dragging details (with the blue area indicating the source region and the red area indicating the target region), the result produced by DragDiffusion\cite{shi2023dragdiffusion}, GoodDrag\cite{zhang2024gooddrag}, Inpaint4Drag\cite{lu2025inpaint4drag}, RegionDrag \cite{regiondrag}, and the result produced by our proposed ICRDrag.
  }
  \label{fig:more_visual_dragbench_dr}
\end{figure*}

\begin{figure*}[t]
  \centering
  \includegraphics[width=.85\textwidth]{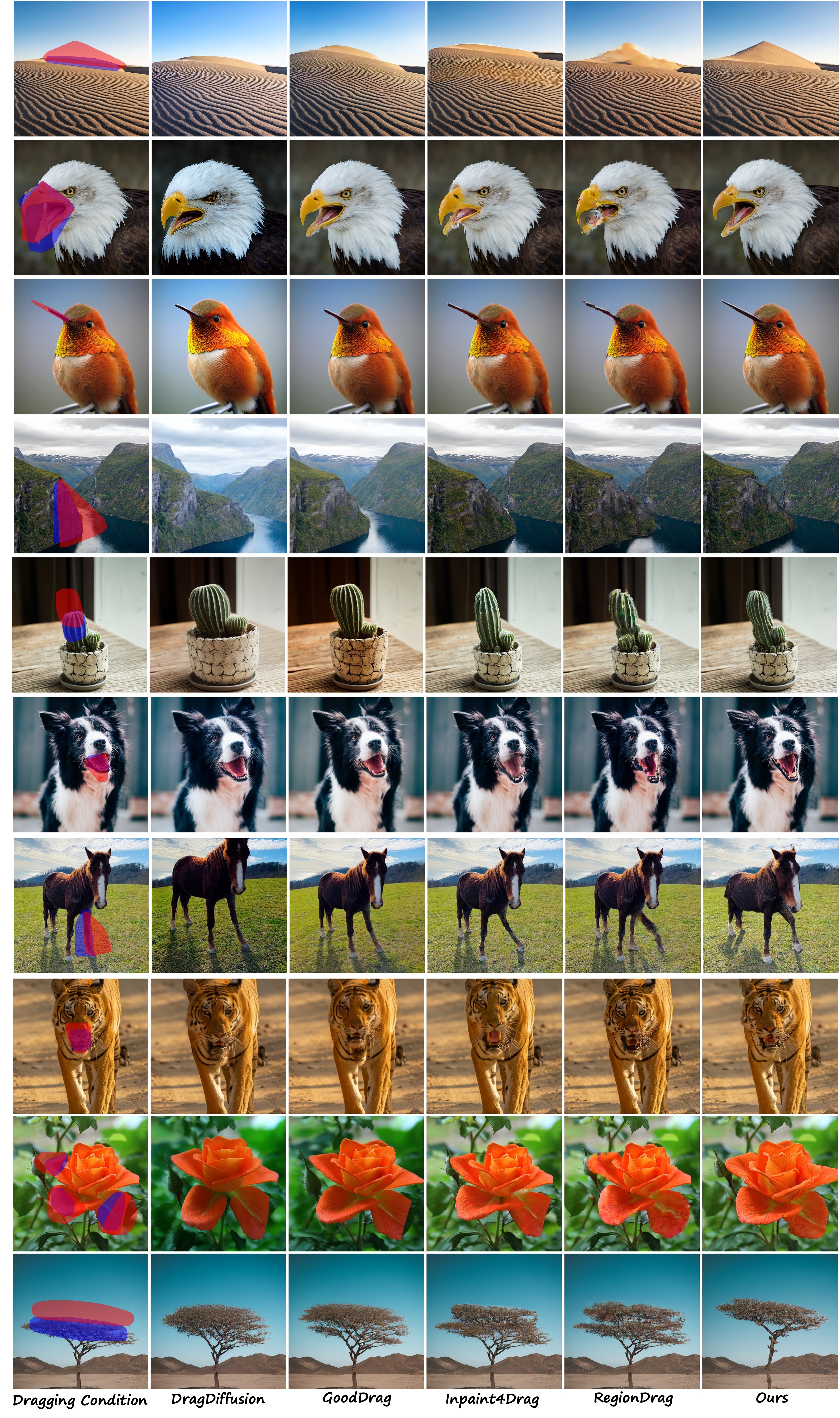}
  \caption{More visual qualitative results on DragBench-SR. In both the left and right subfigures, the visualizations from left to right are as follows: the dragging details (with the blue area indicating the source region and the red area indicating the target region), the result produced by DragDiffusion\cite{shi2023dragdiffusion}, GoodDrag\cite{zhang2024gooddrag}, Inpaint4Drag\cite{lu2025inpaint4drag}, RegionDrag \cite{regiondrag}, and the result produced by our proposed ICRDrag.
  }
  \label{fig:more_visual_dragbench_sr}
\end{figure*}

\begin{figure*} [t]
  \centering
  \includegraphics[width=\textwidth]{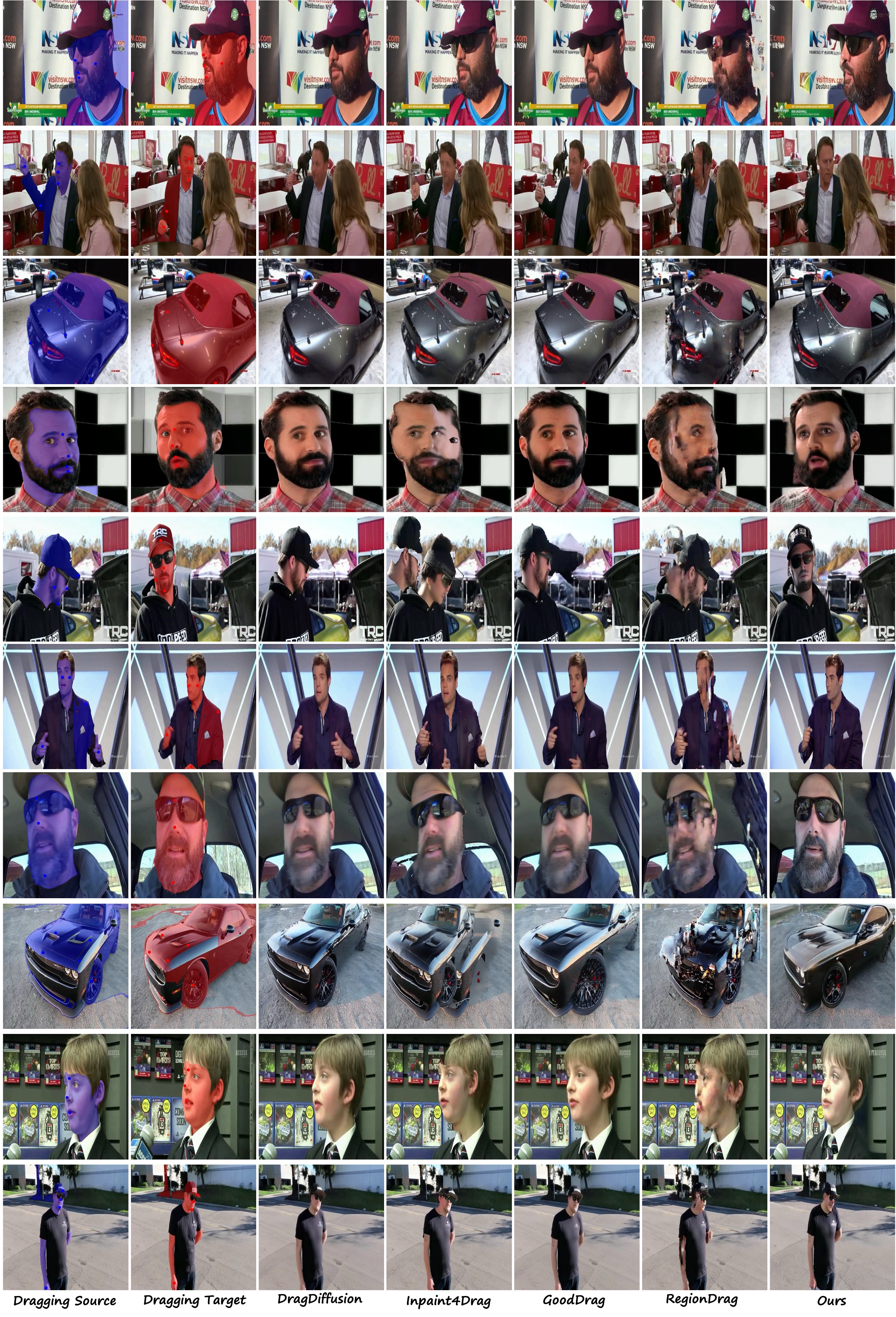}
  \caption{More visual qualitative results on PRD benchmark. The images are arranged from left to right as follows: the source image with the source region highlighted in blue; the target image with the target region highlighted in red; and the result generated by DragDiffusion\cite{shi2023dragdiffusion}, Inpaint4Drag\cite{lu2025inpaint4drag}, GoodDrag\cite{zhang2024gooddrag}, RegionDrag \cite{regiondrag} and our proposed ICRDrag.
  }
  \label{fig:more_visual_prd}
\end{figure*}

In this section, we present additional editing results on both PRD benchmark and DragBench. In Figure \ref{fig:more_visual_prd}, we showcase more examples of our proposed ICRDrag on PRD benchmark. In Figure \ref{fig:more_visual_dragbench_dr} and Figure \ref{fig:more_visual_dragbench_sr}, we provide more comparisons between ICRDrag and the baselines on DragBench. It can be observed that, compared to the baselines, ICRDrag significantly reduces artifacts—such as the altered head orientation of the boy and the deformation of sand dunes. Moreover, ICRDrag better fulfills the intended edits (e.g., repositioning of the tree and the turning of the man’s head), while also preserving fine details more effectively (such as facial features in the head-turning of the boy and the man, and edge details in the adjustments of the table and the sand dunes).
Our proposed ICRDrag achieves strong editing performance even on the broader and more complex images in PRD, and demonstrates superior results compared to the baselines on DragBench.

\section{More Visual Ablations}
\label{sec:ablation}
\begin{figure*} [t]
  \centering
  \includegraphics[width=\textwidth]{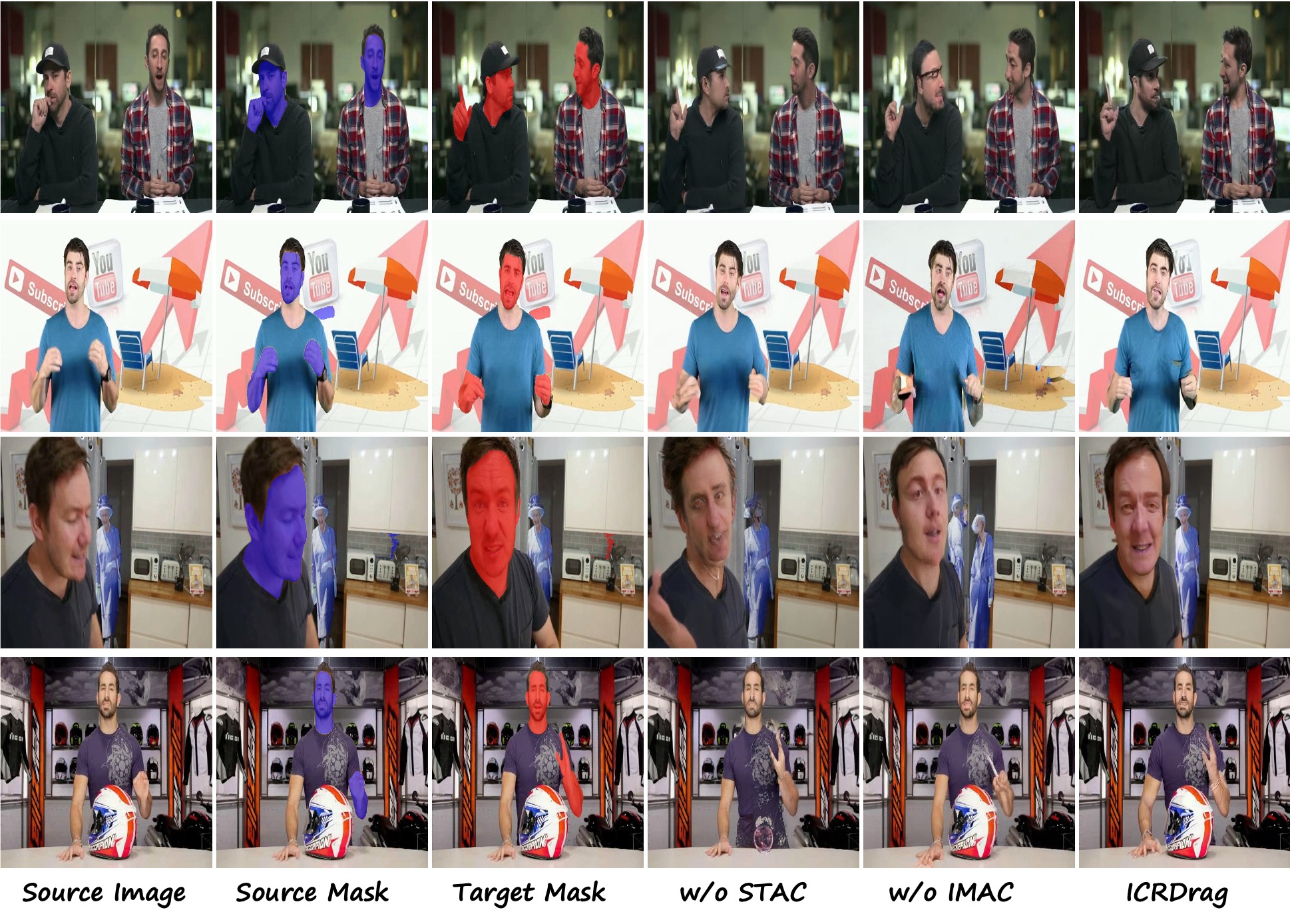}
  \caption{More ablation results of IMAC and STAC losses on PRD benchmark. From left to right, the figure shows the following: the source image; the source image with the source region highlighted in blue; the target image with the target region highlighted in red; the result without STAC loss; the result without IMAC loss; and the result with both losses.
  }
  \label{fig:more_ablation}
\end{figure*}

In Figure \ref{fig:more_ablation}, we present more visual ablation results on the Paired Region Dataset. As can be observed, activating both IMAC and STAC leads to better performance compared to activating either component alone. For example, in the first example, when IMAC is not activated, the alignment between the image and the mask is suboptimal. Conversely, when STAC is disabled, the recovery of fine-grained details in the image deteriorates, suggesting that the attention may be focusing on incorrect source patches.
 When both losses are activated simultaneously, the model can more effectively adjust the man's head orientation based on the region mask, while better preserving the details in the image, and reducing artifacts. This joint activation results in more accurate alignment with the target region mask, improved detail preservation, and reduced artifacts in the edited images.

\section{Failure Cases}
\label{sec:failurecases}

While ICRDrag shows strong performance across various scenarios, its effectiveness can be influenced by the quality of input region masks. In cases where the masks are imprecise or semantically inconsistent, the editing results may contain artifacts. Additionally, performance on certain rare domains (e.g., stylized or animated content) could be further improved with more targeted data. 

\begin{figure*}[t!]
  \centering
  \includegraphics[width=1\textwidth]{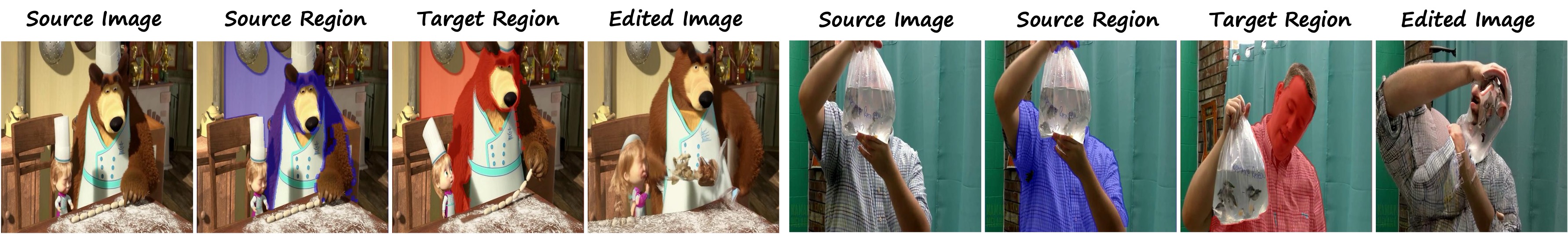}
  \caption{Failure cases of ICRDrag. In both the left and right subfigures, the images are arranged from left to right as follows: the source image; the source image with the source region highlighted in blue; the target image with the target region highlighted in red; and the result generated by our proposed ICRDrag.
  }
  \label{fig:failure_cases}
\end{figure*}

As shown in Figure \ref{fig:failure_cases}, in the first example, although ICRDrag successfully edits the specified region according to the user-provided region masks, it fails to preserve the details in the non-edited areas and introduces noticeable artifacts. This may be attributed to the limited number of similar animated images in the training set. The second failure case stems from an ill-defined region mask. Specifically, while the source region mask includes only the clothing area, the target region mask covers both the clothing and the face, resulting in an inherently inconsistent editing instruction. Consequently, the model fails to generate a plausible image and produces significant artifacts.

\section{Broader impacts}
\label{sec:impacts}

This work advances the capabilities of region-based dragging through a more precise and flexible dragging paradigm. By introducing ICRDrag and the Paired Region Dataset (PRD), we aim to support future research on controllable visual content generation. Potential applications include graphic design, virtual content creation, and assistive editing tools. However, like other generative models, misuse remains a concern—e.g., in creating deceptive or manipulated content. To mitigate this, we encourage responsible use, transparency, and the development of detection tools in downstream applications.

% ---- Bibliography ----
%
% BibTeX users should specify bibliography style 'splncs04'.
% References will then be sorted and formatted in the correct style.
%
\bibliographystyle{splncs04}
\bibliography{main}